\documentclass[final,5p,times,twocolumn]{elsarticle}

\usepackage{amssymb}

\usepackage{lineno}

\usepackage{multirow}
\usepackage{subcaption}
\usepackage{graphicx}
\usepackage{algpseudocode}
\usepackage{algorithm}
\usepackage{amsmath}
\usepackage{verbatim}
\usepackage{amsfonts}
\usepackage{amssymb}
\usepackage{placeins}
\usepackage[normalem]{ulem}
\usepackage{enumerate}
\usepackage{color, soul}
\usepackage{mathtools}
\usepackage{array,arydshln}
\usepackage{url}
\usepackage{titlesec}
\usepackage{graphicx}
\usepackage{mathtools}
\usepackage{algpseudocode}
\usepackage[dvipsnames]{xcolor}

\usepackage{textcomp}
\usepackage{microtype}

\usepackage{booktabs,makecell,tabularx}

\newcolumntype{C}[1]{>{\centering\arraybackslash}p{#1}}
\newcolumntype{L}{>{\raggedright\arraybackslash}X}

\usepackage{siunitx}
\usepackage{etoolbox}

\newrobustcmd{\B}{\bfseries}

\algnewcommand\algorithmicforeach{\textbf{for each}}
\algdef{S}[FOR]{ForEach}[1]{\algorithmicforeach\ #1\ \algorithmicdo}
\DeclareMathOperator*{\argmax}{argmax}

\definecolor{atomictangerine}{rgb}{1.0, 0.6, 0.4}

\journal{Robotics and Autonomous Systems}

\begin{document}
\begin{frontmatter}

\title{Overcome the Fear Of Missing Out: Active Sensing UAV Scanning for Precision Agriculture}

\affiliation[inst1]{organization={Information Technologies Institute, Centre for Research \& Technology Hellas},
addressline={Thessaloniki, 57001},
country={Greece}}
\affiliation[inst2]{organization={Department of Electrical and Computer Engineering, Democritus
University of Thrace},
addressline={ Xanthi, 67100},
country={Greece}}

\author[inst1]{Marios Krestenitis}
\author[inst2,inst1]{Emmanuel K. Raptis}
\author[inst1]{Athanasios Ch. Kapoutsis}
\author[inst1]{Konstantinos Ioannidis}
\author[inst2,inst1]{Elias B. Kosmatopoulos}
\author[inst1]{Stefanos Vrochidis}

\let\oldtwocolumn\twocolumn
\renewcommand\twocolumn[1][]{%
    \oldtwocolumn[{#1}{
    \begin{center}
           \includegraphics[width=0.99\textwidth]{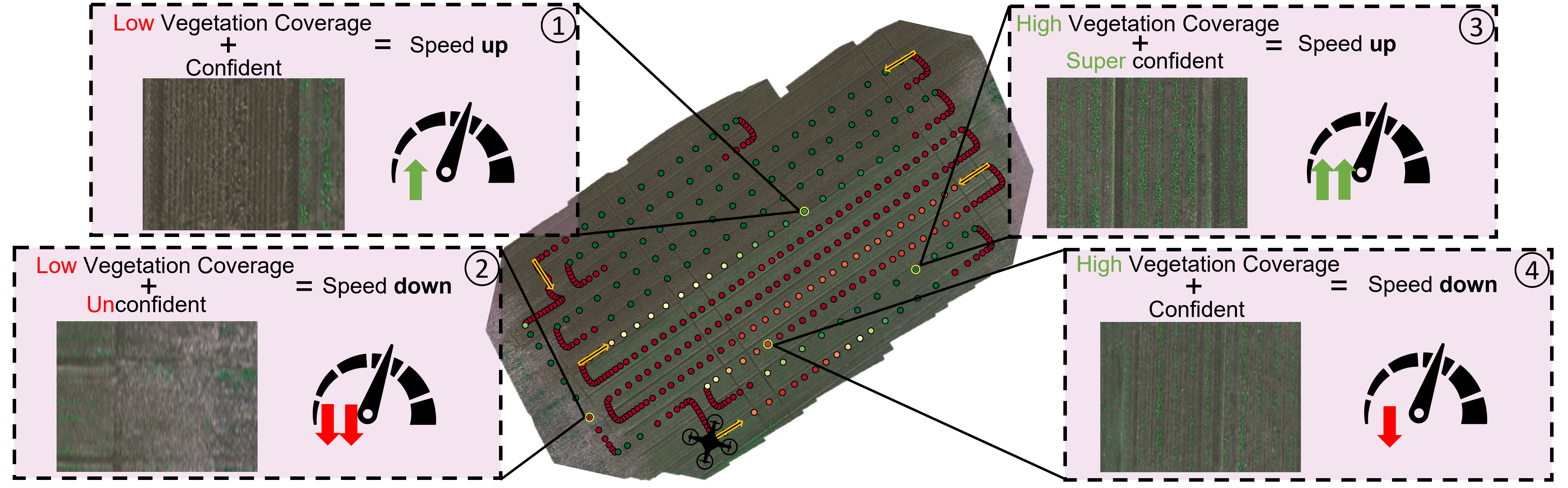}
           \captionof{figure}{Graphical illustration of the core rationale behind the proposed active sensing approach. A UAV is scanning a field with an on board system to estimate the vegetation coverage via captured images, the objective is to on-line regulate its speed so as 1) to cover in detail the whole area and 2) in the minimum possible time. Intuitively, one would like to speed up in areas with little to no information, i.e. vegetation coverage, (Snapshot 1) and slowdown in areas that have rich information to be sure that it can capture everything in great detail (Snapshot 4). However, the amount of information is not the sole factor that should define such changes, as the system that estimates this information could be occasionally inaccurate, mostly due to camera movement. In Snapshot 2, although probably there is not significant information underneath, the UAV should slow down to increase its confidence and be sure about this estimation. On the other hand, Snapshot 3 illustrates a case where, although the vegetation coverage is definitely high, the absolute certainty in such estimation allows for an extra increase in the UAV speed, allowing to save precious flight time.}
           \label{fig:visual_abstract}
        \end{center}
    }]
}

\begin{abstract}
This paper deals with the problem of informative path planning for a UAV deployed for precision agriculture applications. First, we observe that the ``fear of missing out'' data lead to uniform, conservative scanning policies over the whole agricultural field.  Consequently, employing a non-uniform scanning approach can mitigate the expenditure of time in areas with minimal or negligible real value, while ensuring heightened precision in information-dense regions. Turning to the available informative path planning methodologies, we discern that certain methods entail intensive computational requirements, while others necessitate training on an ideal world simulator. To address the aforementioned issues, we propose an active sensing coverage path planning approach, named OverFOMO, that regulates the speed of the UAV in accordance with both the relative quantity of the identified classes, i.e. crops and weeds, and the confidence level of such detections. To identify these instances, a robust Deep Learning segmentation model is deployed. The computational needs of the proposed algorithm are independent of the size of the agricultural field, rendering its applicability on modern UAVs quite straightforward. The proposed algorithm was evaluated with a simu-realistic pipeline, combining data from real UAV missions and the high-fidelity dynamics of AirSim simulator, showcasing its performance improvements over the established state of affairs for this type of missions. An open-source implementation of the algorithm and the evaluation pipeline is also available: \url{https://github.com/emmarapt/OverFOMO}.
\end{abstract}



\begin{keyword}
Adaptive Path Planning \sep Semantic Segmentation \sep UAV Imagery \sep Precision Agriculture \sep AirSim
\end{keyword}

\end{frontmatter}

\section{Introduction}
\label{sec:intro}
    
Unmanned Aerial Vehicles (UAVs) are probably the robotics platforms with the highest adoption rate from professionals in their fields. For example, UAVs are now vital assets for rescuers to quickly search large areas \cite{martinez2021search}, construction engineers to monitor their project's evolution \cite{kim2019remote}, firefighters to quickly assess and identify the fire front \cite{pham2018distributed}, farmers and agronomist to effectively assess the crops health \cite{rodriguez2021assessment}, etc. Such a diverse adoption drives more people and effort to be devoted to UAVs related research and development, leading, in turn, to a further increase in the type of the supported applications. One of the critical factors that have affected this UAV success cycle is the recent advancements in deep learning and specifically in computer vision tasks \cite{o2019deep}. Now, more than ever, we have at our disposal powerful tools that can process the UAV-related data both in an offline and onboard fashion. One of the most severe bottlenecks has to do with the available quantity and quality of data (diverse, clean, and annotated) to deploy the deep learning techniques. Therefore, it is of paramount importance to develop efficient methodologies for automatic meaningful data acquisition, using limited infrastructures.

One of the UAV application areas that could benefit greatly from such methodologies lies within the precision agriculture domain. More precisely, in these applications the UAV collected data are used, in post-processing fashion, to construct homogeneous orthomosaic \cite{gonzalez2020real}, define the crops' health \cite{ampatzidis2020agroview}, find crop line \cite{karatzinis2020towards}, detect and recognize harmful weeds \cite{bah2018deep}, etc. The problem to be investigated in this paper deals with the intelligent design of UAV scanning policy, so as to avoid spending time in areas with little to no real value while being extra precise in information-rich areas. In essence, we seek to answer the following question: Can the online received information ``steer'' the UAV towards a more efficient data collection policy? In literature, this problem is usually referred to as Informative Path Planning (IPP) \cite{ruckin2022informative, meliou2007nonmyopic}.

\subsection{Related Work}
Currently, the vast majority of the UAV agriculture coverage mission planners applies a variance of back-and-forth methodology exploiting the Spanning-Tree Coverage (STC) algorithm \cite{gabriely2001spanning}, or boustrophedon approach \cite{choset1998coverage}. Although this family of approaches is relatively simple, it has been proven quite effective, rendering it the ``go-to'' approach \cite{cabreira2019survey}. The problem with such approaches is the implied assumption of a uniform distribution of the information across the field to be surveyed. In practice, this is rarely the case, forcing the UAV path to be either {\it too pessimistic} and eventually cover fewer square meters than it could or {\it not pessimistic enough} resulting in inadequate coverage in specific subparts of the field. Recognizing that, a fair amount of IPP works have been proposed 
,which deploy a trajectory adjusting mechanism based on the online received data. 

Research-wise, a large number of UAV-based IPP applications have been developed using Gaussian processes (GPs) as a natural way of encoding spatial correlations among the online received data and creating terrain maps of continuous scalar fields. Within the realm of IPP, GPs have gained considerable popularity as a Bayesian method for effectively modeling spatiotemporal phenomena and their inherent correlations \cite{williams2006gaussian}, enabling the collection of data that takes into account both map structure and uncertainty. However, the primary challenge encountered when directly applying Gaussian Processes (GPs) \cite{hitz2017adaptive, vivaldini2019uav} is the significant computational burden that arises due to the accumulation of dense imagery data over time.

Ruckin et al. \cite{ruckin2022informative} introduced an IPP methodology utilizing Bayesian techniques as an active learning acquisition function to quantify the pixel-wise model uncertainty in semantic segmentation. Their approach aimed to maximize the improvement in the model's performance by assimilating the most informative terrain data with the highest uncertainty, linking thus the information gain from the active learning acquisition function to a planning objective. Vivaldini et al. \cite{vivaldini2019uav} proposed an online UAV-based IPP system, wherein the acquisition function is designed to minimize the uncertainty associated with the differentiation between diseased trees and healthy trees as well as roads in a Gaussian map interpolation. The path planning module strategically selects sampling points to achieve comprehensive environmental coverage, utilizing the Rapidly-exploring Random Trees (RTT) algorithm to optimize the gathering of crucial information. To minimize the distance traveled and ensure sufficient coverage of the surveyed area, an objective function is responsible for guiding the UAV toward reducing the average uncertainty of an image at a given position (x, y) on the current classification map. Although their experimental evaluation demonstrated favorable outcomes in comparison to static coverage paths, the proposed methodology allocates the UAV's battery life to repetitive back-and-forth movements, which undeniably leads to suboptimal efficiency in continuous terrain monitoring.

Popovic et al. \cite{popovic2017multiresolution, popovic2020informative} proposed an IPP framework for active classification, exploiting the spatial correlation encoded in a Gaussian Process model as a prior for Bayesian data fusion to facilitate expedited map updates. They proposed an adaptable path-planning approach that generates dynamically viable trajectories at varying altitudes in a continuous 3D space to achieve high-quality aerial imaging with constant-time measurements by computing the informative objective with the new map representation. Their strategy, however,  assumes swift map updates with minimal computational overhead, while simultaneously allocating the UAV's temporal resources to vertical maneuvers. While their simulated and real-life experiments yielded positive results when compared to static coverage paths, it is noteworthy that the suggested methodology has predominantly been appraised in limited-scale field trials where the temporal exigency of the UAV's battery life is relatively inconsequential. This attribute assumes critical significance, particularly in vast spatial domains, as the allocation of the UAV's battery life to the monitoring of new informational content becomes an overriding concern. In contradistinction, our study employs real-time sensor data and progressively generates adaptable speed-based trajectories at a continuous pace over time, wherein the computational demands for online recalculations remain decoupled from the temporal prerequisites for map revisions, as they solely rely on the present image acquisition.

Stache et al. \cite{stache2021adaptive} proposed an IPP framework for precision agriculture, specifically targeting crop/weed segmentation, similar to our work. The distinguishing characteristic of their methodology lies in the incorporation of an accuracy model for deep learning-based architectures, enabling the quantification of the relationship between UAV altitude and semantic segmentation accuracy. They introduced a dynamic path planning approach based upon the boustrophedon method within a continuous 3D spatial domain, generating evolving trajectories at various altitudes for monitoring and close inspection tasks. Nevertheless, their approach, which involves replanning at variable altitudes, prioritizes the acquisition of higher-resolution data over minimizing flight time for comprehensive monitoring of the entire agricultural field.

One of the major factors that hinder the wider appliance of such methods is the computation needs during each replanning phase. Additionally, because several candidate paths along with their anticipated measurements should be simulated before each replanning step, their computational needs grow exponentially with respect to the field area to be covered. Previous studies have developed quite elaborate plans to mitigate this by pruning the action-space \cite{popovic2020informative, koutras2020autonomous}; however, this kind of relaxation could seriously degrade the quality of the achieved performance. Recent approaches attempt to mitigate this issue by treating the IPP as a standard Reinforcement Learning (RL) problem, learning policies that are able to compute inexpensive plans online \cite{ruckin2021adaptive}. However, the performance of this approach is highly correlated to the matching between the real world and the simulative environment with realistic data that the RL agent will be trained on. Last but not least, the majority of the available approach does not incorporate a hard constraint with respect to the available battery of the UAV, rendering their realization particularly tricky.

Aiming to overcome this ``fear of missing out'' important data, we propose OverFOMO, an active sensing coverage path planning approach that adopts the STC algorithm as a blueprint for the UAV path while, depending on the online received information, it adjusts its focus on specific areas. Assuming an UAV covering an agricultural field, figure \ref{fig:visual_abstract} illustrates the proposed active sensing approach using 4 key snapshots. Snapshots 1 and 4 depict two representative examples that define the core motivation behind the proposed system, revealing that the quality of received information is inversely proportional to the UAV’s speed, basically due to blurring effects. More specifically, the received image in snapshot 1 contains only a few crops and is relatively clear; therefore, the UAV can afford to speed up. Snapshot 4 presents a case where the received image is full of vegetation, but the speed of the UAV makes the segmentation process less confident. In that case (Snapshot 4), the UAV should slow down to make more accurate detections, especially in this high vegetation density subpart of the field. Snapshots 2 and 3 present the ability of the proposed active sensing scheme to handle ``tricky’’ cases. Snapshot 2 seems to contain low vegetation coverage; however, the predicted segmentation is insecure, and therefore the UAV should speed down rapidly to verify that indeed there are no missing crops around that area. Moving to the other side of the spectrum, the received image in snapshot 3 contains much vegetation; however, the on-board segmentation process is super confident about the identification and therefore, the speed can be safely increased without sacrificing loss of information.

Although there have been proposed several alternatives to the usual practice with the back-and-forth movements \cite{gabriely2001spanning, choset1998coverage}, that online calculate the next monitoring position (e.g., previously mentioned IPP methods), their time efficiency is usually significantly reduced \cite{kapoutsis2016real, renzaglia2012multi}, leaving the back-and-forth movements as the “go-to” option for this type of missions. Within this paper, instead of proposing another approach that calculates the best next monitoring position, we strategically combine elements of the two approaches to achieve beyond state-of-the-art performance. More specifically, we keep the back-and-forth movements as the blueprint of the UAV path to also retain the performance guarantees that come with such an approach, and, at the same time, we attempt to regulate online the time spent in each sub-area based on the local information, similar to what a person would do. In a nutshell, the contributions of this work are:
\begin{itemize}
    \item Development of a novel active sensing coverage path planning scheme that inherits the STC optimality and completeness guarantees. The computational needs for the online recalculation do not depend on the size of the operation field, making it suitable for various applications while respecting operational constraints (e.g., remaining battery, etc.).

    \item Development of a novel Deep-Learning-based module for adjusting the UAV speed, similar to what a human would do, taking into consideration both the quantity of the detected relevant instances  (i.e. crops and weeds) and certainty (quality) about these detections. Note that the method could be extended to different operational scenarios, e.g. scan a sea area and regulate UAV speed according to marine-related classes (oil spill, algae bloom, etc.)    

    \item An open-source, modular, simurealistic pipeline that combines the high-fidelity dynamics of AirSim \cite{shah2018airsim} with real RGB images sourced from publicly available UAV datasets.    
      
    \item Validation of the proposed approach, using the aforementioned simurealistic pipeline, against the widely-used STC-based coverage methods, showcasing its performance. Contrary to the prevailing state-of-the-art STC planners \cite{karatzinis2020towards, pham2017aerial}, which entails a uniform scanning speed across the surveyed region, our method incorporates adaptive agent speed, resulting in reduced flight duration and enhanced image quality.

\end{itemize}

\section{Problem formulation}
\label{sec:problem_formulation}
Following the standard UAV-based monitoring Precision Agriculture (PA) process \cite{2018weedmap, krestenitis2022cofly, radoglou2020compilation}, we assume a UAV capable of acquiring images mounted with an RGB camera flying at fixed altitude. The studied problem is defined as controlling in-real-time the UAV mission parameters i.e speed, so as to acquire the best possible field representation, i.e the fidelity of field orthomosaic, within the minimum flight time.

\subsection{Decision Variables}
\label{subsec:decision_variables}
Assuming a fixed sampling rate, i.e.  images per time, the IPP setup is reduced to design the series of sensing waypoints that will comprise the UAV trajectory:
\begin{equation}
\label{eq:trajectory}
    \tau = \left[ w_1, w_2, \dots, w_n\right],
\end{equation}
where $w_i \in \mathbb{R}^2$ denotes the image capturing position in the plane of the operational height. The time needed to complete a trajectory is denoted with $C(\tau)$ and it should be less or equal to the maximum operational flight time $T_{max}$ of the UAV.

\subsection{Field Representation Quality Assessment}
\label{subsec:objective}
After the completion of the UAV mission, all $\left\lbrace I_1, I_2, \dots, I_n \right\rbrace $ images, gathered from the sampling positions as defined in (\ref{eq:trajectory}), are going to be stitched to generate the field's orthomosaic. The quality of the extracted orthomosaic is related to the quality of the captured images $\left\lbrace I_1, I_2, \dots, I_n \right\rbrace $ and proportional to the comprehensibility of the enclosed semantic content. 

A common approach to measure this attribute is segmenting relative instances over the generated orthomosaic, such as crops and weeds from soil \cite{krestenitis2022cofly}. The discrimination capability of a well-trained and robust segmentation model is related to the quality of the visual input. As in the majority of semantic segmentation problems, the model efficiency gets assessed by using the Intersection over Union (IoU) \cite{rezatofighi2019generalized}.

\subsection{Informative Path Planning Problem}
\label{subsec:ipp_problem}
Having defined the estimation approach for the field's representation quality, the general IPP problem, under the context of precision agriculture, can be translated to the following optimization problem:

\begin{equation}
\label{eq:OptimizationProblem}
    \begin{aligned}
        & \underset{\tau}{\text{maximize}}
        & & \frac{\mbox{IoU}^{\mbox{\textit{crop}}}(\tau) +\mbox{IoU}^{\mbox{\textit{weed}}}(\tau)}{\alpha C(\tau)} \\
        & \text{subject to}
        & & C(\tau) - T_{max} \leq 0
    \end{aligned}
\end{equation}
where $\alpha$ is used to weight $C(\tau)$ in terms of $\mbox{IoU}^{\mbox{\textit{crop}}}(\tau) +\mbox{IoU}^{\mbox{\textit{weed}}}(\tau)$, depending on the specifics of each application. For example, a usual configuration is targeting for the best possible representation (terms: $\mbox{IoU}^{\mbox{\textit{crop}}}(\tau) +\mbox{IoU}^{\mbox{\textit{weed}}}(\tau)$) within a given time budget $T_{max}$. For this configuration, $\alpha$ is chosen to be appropriate small to render the influence of the denominator technically negligible (of course, the constraint always holds). 

A direct difficulty in solving (\ref{eq:OptimizationProblem}) lies within the immense continuous domain of (\ref{eq:trajectory}). Actually, the number of different possible combinations of (\ref{eq:trajectory}) increases exponentially with respect to the size of the field \cite{galceran2013survey}, which determines the number  $n$ of image capturing positions. However, the most severe obstacle has to do with the fact that both the explicit forms of $\mbox{IoU}^{\mbox{\textit{crop}}}(\cdot)$ and $\mbox{IoU}^{\mbox{\textit{weed}}}(\cdot)$ are not available prior to the UAV mission, since ground-truth information is required. As a consequence, any approach that relies on evaluating different combinations of (\ref{eq:trajectory}) on (\ref{eq:OptimizationProblem}) cannot be realized within this context. On the contrary, the solution should be seeked in a method capable to assess during the ongoing mission the quality of the captured images and regulate the UAV speed accordingly, in order to extract the best possible field representation in the minimum flight time. Toward this direction, one of our main objectives is to deploy an approach that tackles (\ref{eq:OptimizationProblem}) and its limitations, in a indirect manner.

\section{Adaptive Coverage Path Planning}
\label{sec:adaptiveCPP}
This section describes the details of OverFOMO, the proposed active sensing coverage path-planning algorithm, designed for previously defined optimization problem (\ref{eq:OptimizationProblem}). 

\subsection{Problem Translation using Coverage Path Planning}
\label{subsec:problemTranslation}
First, let us define the Coverage Path Planning (CPP) problem \cite{cabreira2019survey} that is defined by the geometry of the agricultural field and the UAV characteristics. In short, Coverage Path Planning problem deals with the problem of designing a robot path that covers an area of interest in the minimum possible time. One of the most popular CPP approaches is Spanning-Tree Coverage (STC) \cite{gabriely2001spanning} algorithm. STC first discretizes the operational area and then generates a minimum spanning-tree that will be used as a guide for the robot path. Overall, STC algorithm is a polynomial time algorithm, with respect to the field size, that guarantees complete grid coverage in the minimum possible time \cite{gabriely2001spanning}. Hence, STC algorithm can be realized as a kernel for optimal coverage paths and generate a sequence of sensing waypoints (\ref{eq:trajectory}), as follows:
\begin{equation}
    \label{eq:stcKernel}
    x = \mbox{STC}({\cal{P}}, o, h, dt, s)
\end{equation}
where ${\cal P}$ denotes the polygon that contains the agricultural field, $o$ is the overlap between two images in adjacent flight path lines \cite{apostolidis2022cooperative}, $h$ is the UAV flight altitude, $dt$ denotes the time-lapse interval for the capture of each image and $s$ is the UAV speed. 

Due to the fact that we are dealing with the IPP for a specific field, ${\cal P}$ is considered known and constant. $o$ and $h$ are defined according to the specifics of each agricultural mission, e.g., plant growth rate, season, required resolution of the orthomosaic, etc. The remaining two parameters are the ones that dominate the density of the captured images ($I_i \mbox{ from } w_i$ position) along the STC-based path. The list of STC parameters can be further reduced, by setting the $dt$ to its smallest feasible value for the onboard sensor that does not compromise the quality of the received images. After these realizations, STC-based trajectory for a given agricultural field and a given type of UAV can be defined as:
\begin{equation}
    \label{eq:stcReduced}
    x = \mbox{STC}(s)
\end{equation}

Hence, utilizing (\ref{eq:stcReduced}), we now have a UAV path that completely covers the agricultural field at the minimum possible time for a given $s$. Inevitably, the definition of $s$ gives rise to a trade-off. A small $s$ value would provide premier quality on the captured images and, therefore, in our ability to distinguish accurately crops and weeds, however, it would result in covering only a small fraction of the agricultural field, due to flight time limitations. On the other hand, an increased $s$ value could mitigate this by covering larger areas, in the expense of our discrimination accuracy. The usual practice is to apply a constant $s$ at the beginning of the mission and perform the whole mission with such speed \cite{karatzinis2020towards, 2018weedmap, pham2017aerial, krestenitis2022cofly}. However, during the operation, the UAV receives images that characterize the quantity of useful information that lies under its current path.

 Within this paper, we want to exploit this online-received information and adjust the speed of the UAV during its flight, making the data acquisition process more efficient. Thus, assuming that $t_i- t_{i-1}$ denotes a fixed time-interval needed for both the information assessment and the change in the UAV speed, we want to guide the image capturing process by the following adaptive path-planning scheme:

	\begin{equation}
		\label{eq:adaptive_path}
		x(t)= 
		\begin{cases}
			\mbox{STC}(s_1),&  t_0<t\leq t_1\\
			\mbox{STC}(s_2),&  t_1<t\leq t_2\\
			& \vdots \\
			\mbox{STC}(s_n),&  t_{n-1}<t\leq C(\tau)\\
		\end{cases}
	\end{equation}

Hence, by plugging the time-varying $x(t)$ into $\tau$, the optimization problem of (\ref{eq:OptimizationProblem}) now is reduced to online adjust $s$ for every time-interval of (\ref{eq:adaptive_path}). In the upcoming subsections, we discuss the details of speed adjustment, i.e. calculating online $\{s_1, s_2, \dots, s_n\}$ of (\ref{eq:adaptive_path}), based on the online-received information gain at each time-interval.

\subsection{Coverage Ratio \& Confidence Level}
\label{subsec:assessIG}
Before providing the exact methodology that online adjusts the speed of the UAV, let us first define two key metrics that assess the information gain with respect to
the current image frame that corresponds to the $i$-th time-interval in (\ref{eq:adaptive_path}).

\begin{figure*}[htpb]
    \centering
    \includegraphics[width=0.99\textwidth]{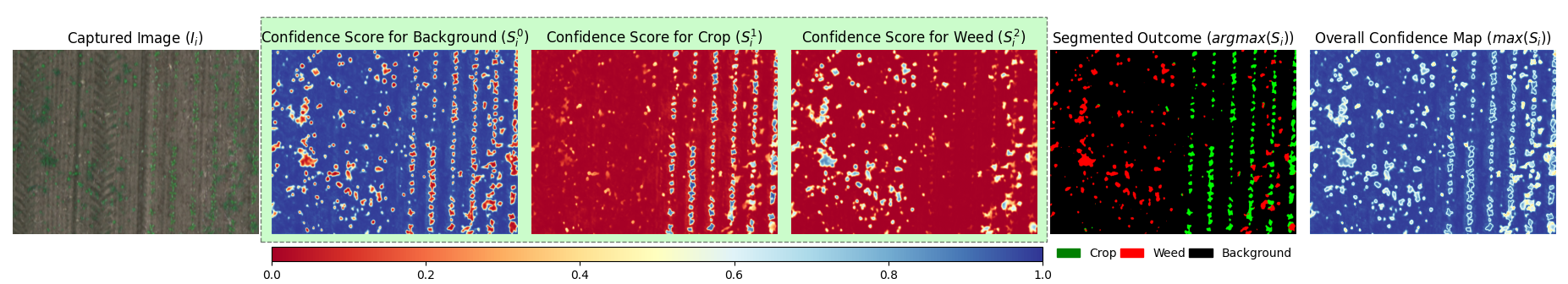}
    \caption{Demonstration example of the semantic segmentation model output which is employed to calculate $cr$ and $cl$ metrics. Captured image $I_i$ is fed to the model and produces the 3-channel array $S_i$ (illustrated per channel and highlighted with pale-green color), where each channel contains the probability of the image pixels belonging to the corresponding class.  $S_i^{class} = \argmax \left( S_i \right)$ leads to the segmented outcome, where each pixel is assigned to one of the $3$ classes, enabling the estimation of $cr$ metric. The overall confidence map, providing the probability of each pixel belonging to the assigned class, is acquired via $S_i^{prob} =\max(S_i)$ and enables the calculation of $cl$ metric.}
    \label{fig:s_array_example}
\end{figure*}

The main rationale is the fact that the information enclosed in the captured $I_i$ image is correlated to the amount of depicted crops and weeds. Thus, a deep-learning model ${\cal M}$ capable of semantically segmenting images to identify three classes, namely crop, weed and background, is deployed. ${\cal M}$ is fed with the acquired $w \times l$ image $I_i$ and produces a confidence score map $S_i \in \mathbb{R}^{w \times l \times 3}$ that contains the probability of each pixel belonging in each class, i.e. $S_i = {\cal M}(I_i)$. As a direct outcome, the prediction mask is derived using $S_i^{class} = \argmax \left( S_i \right)$, assigning a class id for every pixel. While $S_i^{prob} =\max(S_i)$ derives the overall confidence map, containing the probability of each pixel belonging to the assigned class. For improved clarity, a visual representation of the aforementioned terms is provided in figure \ref{fig:s_array_example}.

The first metric is oriented to quantify the amount of the captured crops and weeds. To accomplish that, we utilize the \textit{coverage ratio} ($cr$), inspired by \cite{stache2021adaptive} and defined as follows:
\begin{equation}
    \label{eq:coverageRatio}
    cr(S_i^{class}) = \frac{N^{crop} + N^{weed}}{N^{crop} + N^{weed} + N^{background}}
\end{equation}
where $N^{crop}$, $N^{weed}$ and $N^{background}$ denotes the number of pixels from $S_i^{class}$ that have been classified in each class correspondingly. Note that the denominator resembles the total number of pixels in the image frame. Conceptually, $cr:\mathbb{R}^2 \rightarrow [0, 1]$ estimates the plants and weeds coverage on the target area by applying (\ref{eq:coverageRatio}) rule in a pixel-wise segmented image of $I_i$. Low values of $cr(S_i^{class})$ imply that the vegetation enclosed in the captured image is limited and thus, the information gain of this area is low. Correspondingly, high values of $cr$ are related to areas of lush vegetation, where the information gain is considered high. 

Having this in mind, a simple formulation for the speed in (\ref{eq:adaptive_path}) would be a linear mapping between $cr$ and the speed, i.e. as the $cr$ is increased the speed gets decreased and vice versa. However, such a formulation can have several pitfalls since the detected weed/crop instances' accuracy is not considered.

Towards this direction, the confidence of the acquired predictions is included as a second metric for assessing the information gain. More specifically, for every processed image $I_i$ the \textit{confidence level} ($cl$) is calculated as follows:
\begin{equation}
    \label{eq:confidenceLevel}
    cl(S_i^{prob},~S_i^{class}) = \frac{\sum_{j \in {\cal C}}{p_j} + \sum_{j \in {\cal W}}{p_j}}{N^{crop} + N^{weed}}
\end{equation}
where ${\cal C}$ and ${\cal W}$ denote the set of pixels that have been annotated as crop and weed, respectively, and $p_j$ denotes the corresponding confidence score for $j$-th pixel of $S_i^{prob}$.

The main idea here is that when the confidence level $cl: \mathbb{R}^2 \rightarrow [0, 1]$ of the acquired prediction is high enough, then the UAV speed can be increased to reduce the flight time since the captured image quality is adequate to make robust predictions. Respectively, a lower confidence level may imply that the quality of the processed image is low, and thus, the UAV should decrease its speed to capture a clearer view of the scene. With respect to the information gain, the confidence level can be considered as an inverse metric of the observed entropy. Higher values imply that the scene is well-known to the prediction model ${\cal M}$ and it can be clearly conceived; thus, the UAV can proceed faster since the acquired information is limited in this {\it static} environment. On the contrary, lower confidence level values imply an {\it unknown} environment, conceived with ambiguity; thus, speed should be decreased to increase the observation time.

\subsection{Speed Adjustment}
\label{subsec:speedAdjustment}
\begin{figure}[h]
    \centering
    \includegraphics[width=0.99\columnwidth]{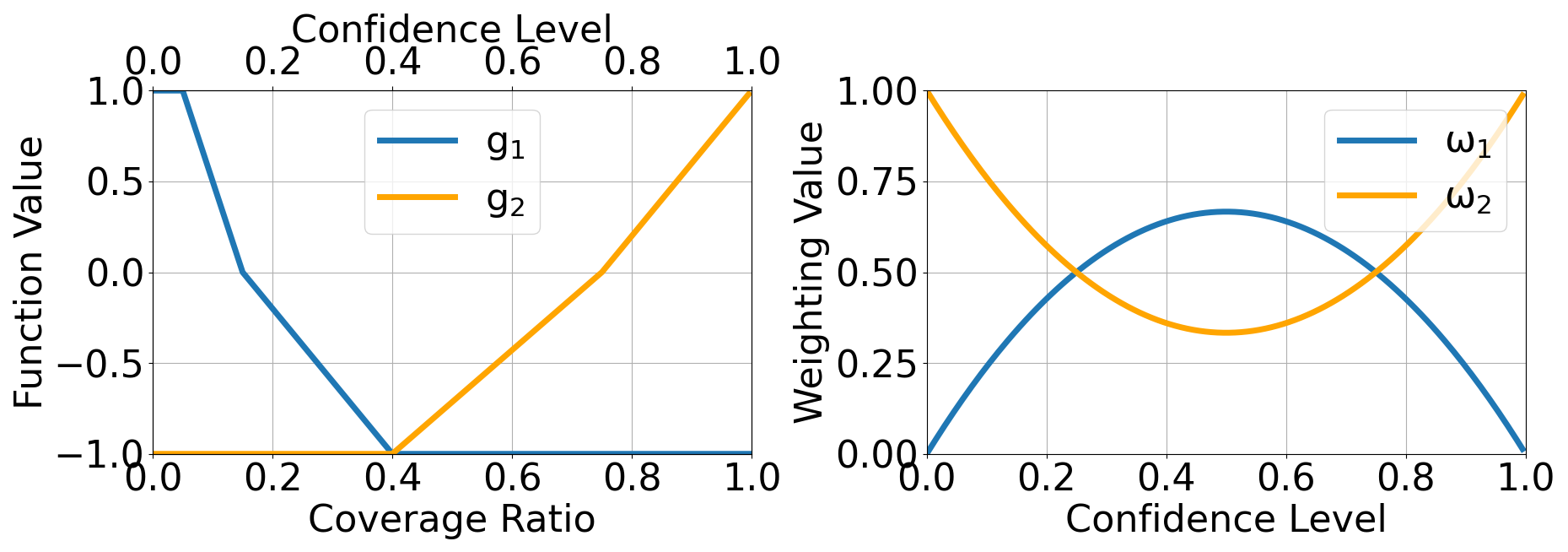}
    \caption{Graphical illustration of the employed functions in (\ref{eq:changeUAVspeed}). Translation functions (left) $g_1(\cdot)$ and $g_2(\cdot)$ aim to map the calculated $cr$ and $cl$, respectively, to a relative speed change. Weighting functions (right) $\omega_1(\cdot)$ and $\omega_2(\cdot)$ aim to regularize the contribution of $g_1(\cdot)$ and $g_2(\cdot)$ to the final decision.}
    \label{fig:functionsWeights}
\end{figure}

Having calculated $cr$ (\ref{eq:coverageRatio}) and $cl$ (\ref{eq:confidenceLevel}) for the currently received $i$-th image, we can now calculate the objective speed adjustment. To perform this update we need a mapping function $G(\cdot):\mathbb{R}^2 \rightarrow [-1, 1]$ that translates both $cr$ and $cl$ into speed changes with respect to the maximum allowed discrepancy $q$ around UAV's nominal speed $\bar{s}$, i.e. 

\begin{equation}
    \label{eq:speedUpdate}
    \begin{split}
    s_{i} &= \mbox{clip} \left(u,\; \bar{s}-q,\;  \bar{s}+q\right),  \\
    u &= s_{i-1} + G(cr, cf)q, ~\mbox{with } s_0 = \bar{s}
    \end{split}
\end{equation}
where clip function constrains the updated speed between safe/acceptable bounds. Hence, to derive the needed behavior in terms of speed change, $G(\cdot)$ is defined as follows:

\begin{equation}
    \label{eq:changeUAVspeed}
    G(cr, cf) = \omega_1(cl)g_1(cr) + \omega_2(cl)g_2(cl)
\end{equation}
where $g_1(\cdot)$ and $g_2(\cdot)$ denote the translation functions from $cr$ and $cl$, respectively, to a relative speed change. Additionally, for each term a regulation function is defined, namely $\omega_1(\cdot)$ and $\omega_2(\cdot)$, to prioritize one term over the other. $g_1(\cdot)$ and $g_2(\cdot)$ have chosen to be linear piecewise functions, while $\omega_1(\cdot)$ and $\omega_2(\cdot)$ are of type of parabola with respect to their parameters. For ease of understanding, figure \ref{fig:functionsWeights} graphically illustrates the form of these functions. Additional information regarding the calibration of $g(\cdot)$ and $w(\cdot)$ functions is provided in \ref{AppendixB}.

Note that the speed adjustment $s_{i}$ in position $w_i$ is with respect to the previous speed ``state'' $s_{i-1}$, instead of the nominal speed $\bar{s}$. The specific choice enables more smooth transitions of the vehicle speed, while the adapting process can be considered to some extent stateful. Furthermore, contrary to the established approaches, the presented method does not adjust the overlap among consecutively captured images to a fixed value \cite{radoglou2020compilation}. To this end, tuning parameter $q$ is enabled to regulate the range of the speed adjustments and, thus, maintain the image overlap within acceptable (application-wise) thresholds \cite{tsouros2019review}. Towards this direction, the proposed method aims to control the quality of the captured image data by adjusting the vehicle speed (with respect to the semantic content of the scene) and, thus, regulating the image distortion due to motion blurring. Both $q$ and $\bar{s}$ are user-defined parameters that can express both the user requirement and the UAV hardware characteristics.

\begin{figure}[htpb]
    \centering
    \includegraphics[width=0.95\columnwidth]{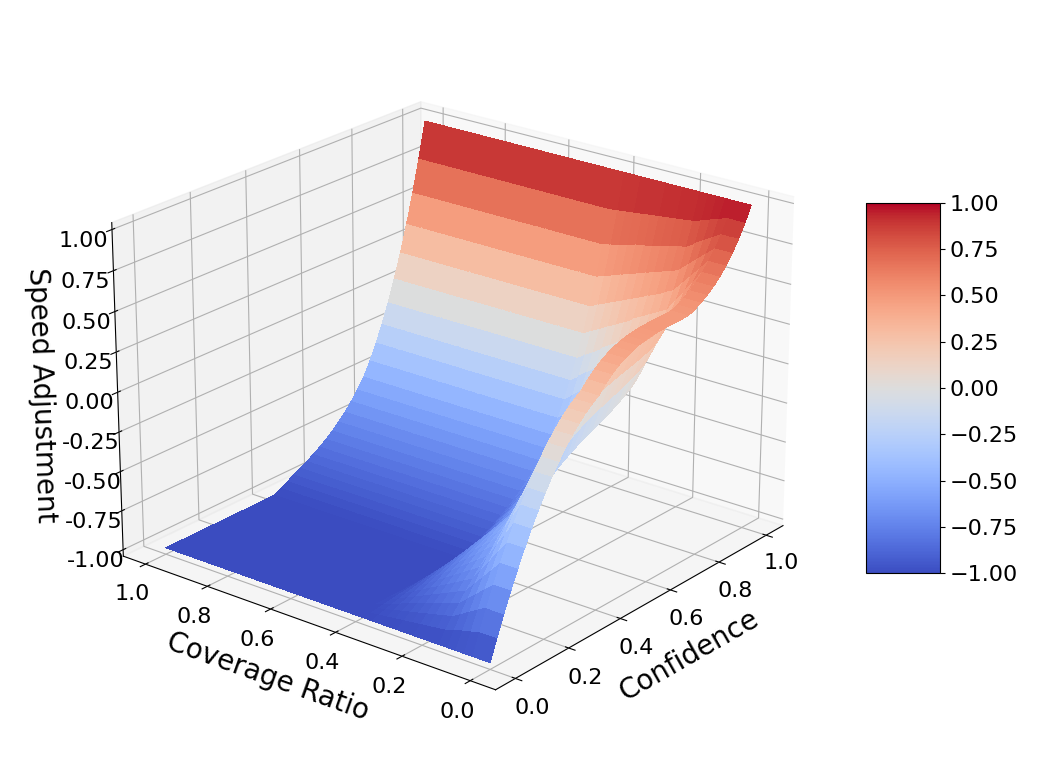}
    \caption{3D graph of the designed $G(\cdot)$ function to adapt UAV speed according to the information gain.}
    \label{fig:g_func_3d}
\end{figure}

The main rationale of weighting functions $\omega_1(\cdot)$ and $\omega_2(\cdot)$ in (\ref{eq:changeUAVspeed}) is to adjust the contribution of each term based on the confidence of the prediction. For instance, assuming that $cl$ value is $0$, then the estimated value of $cr$ and, by extension the value of $g_1(cr)$ is irrelevant since it is based on inaccurate predictions. Similarly, in the case of $cl = 0.5$ the prediction can be considered to some extent as ambiguous and the formulation favors coverage ratio measurements in order to regulate speed\footnote{Please note that, although the information gain can be described quite effectively by these functions, their forms can be further fine-tuned to achieve better, problem-oriented performance.}. Aiming to provide further insights regarding the system's behavior under different scenarios, in figure \ref{fig:g_func_3d} is demonstrated a 3D representation of $G(\cdot)$ function for its whole domain.

\subsection{Proposed Method as a Whole}

Having analyzed the key points in the previous sections, the proposed adaptive path planning can be summarized as ``estimate the information gain captured in image $I_i$ and adapt the vehicle speed according to it''. Since there is no ground truth, we employ the two metrics, coverage ratio ($cr$) and confidence level ($cl$), in order to tackle its absence and concurrently quantify the information enclosed in each image. Coverage ratio estimates the amount of crops and weeds in the scene and aims to answer the question ``how much significant is this area?''. Confidence level aims to quantify the validity of the model estimation and responds to the question ``how much accurate though is the estimation regarding the significance of this specific area?''. 

\begin{figure*}[ht!]
    \centering
    \includegraphics[width=0.99\textwidth]{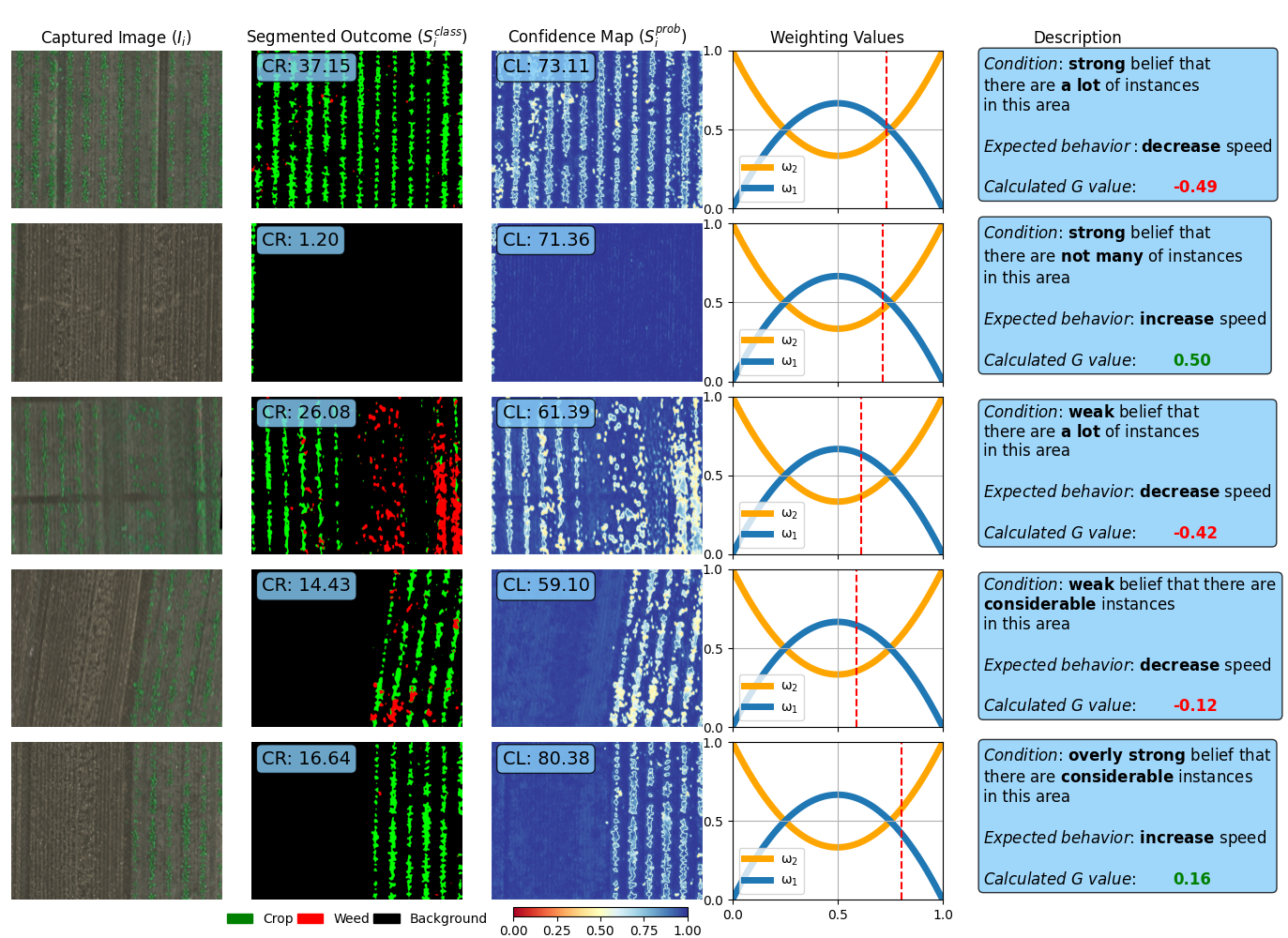}
    \caption{Illustration of different operational cases of the adaptive coverage path planning. Each row presents the analysis conducted for the corresponding captured image. For each case, the corresponding $cr$ and $cl$ metrics are mentioned (\%). Vertical red line in the figures of the fourth column corresponds to the $cl$ metric, based on which the weighting values are calculated and employed in (\ref{eq:changeUAVspeed}) to calculate the corresponding $G$ value. Last column provides a short description of each case among with the expected behavior of an adaptive system and the corresponding $G$ value of our method, which is employed in (\ref{eq:speedUpdate}) to update the UAV speed accordingly.}
    \label{fig:different_cases}
\end{figure*}

At each step, the proposed method answers these two questions and regulates the UAV speed accordingly through function $G(\cdot)$. In figure \ref{fig:different_cases} we present a comprehensive set of operational scenarios, providing insights regarding the expected behavior of an adaptive system that self-regulates its speed, which was our main motivation, along with the key-values of the OverFOMO that lead to the corresponding adjustment. The first two rows of the figure refer to cases where the model can provide a concrete estimation regarding the amount of existing crops or weeds and the UAV speed is regulated according to the quantity of the detected instances. Rows 3 and 4 refer to cases where the quality of captured data deteriorated due to motion blurring. One can notice the impact of this effect on the calculated $cl$ metric. Despite the amount of estimated crops and weeds, the vehicle speed is decreased since the quality of captured data implies ambiguous estimations. The last row resembles the case where the estimator is overly confident implying that the data quality is adequate and therefore a partial deterioration, by increasing the UAV speed, can be tolerated to save flight time. As demonstrated, the proposed adaptive scheme can confront variable cases. In this direction, the proposed adaptive scheme considers the quantity ($cr$) and the quality ($cl$) of the information gained per image, aiming to operate in a sweet spot where the quality of captured data is maximized while the flight time is minimized.

In a nutshell, Algorithm \ref{alg:algorithm} outlines the proposed adaptive coverage path planning as a whole. Putting everything together, the proposed approach alleviates both the combinatory nature and the unknown factor by a careful combination of two ingredients: i) the STC algorithm that is capable of computing offline optimal coverage paths with ${\cal O}(n)$ complexity, and ii) an online speed adjustment scheme that takes into consideration the current information gain. 

\begin{algorithm}[ht!]
    \caption{Adaptive Coverage Path Planning}\label{alg:algorithm}
    \begin{algorithmic} [1]
        \Require ${\cal{P}}, o, h, dt, \bar{s}, q, {\cal M}$
        \Ensure $\tau$
        \Statex {\it Offline phase}: 
        \State Define a STC-based trajectory parametric over $s$ (\ref{eq:adaptive_path})
        \Statex {\it Online phase}:
        \ForEach {viewpoint $w_i$ at $t_i$}
        \State Acquire frame $I_i$
        \State $S_i \leftarrow {\cal M}(I_i)$ and $S_i^{class} \leftarrow \argmax \left( S_i \right)$
        \State Calculate $cr$ and $cl$ according to (\ref{eq:coverageRatio}) and (\ref{eq:confidenceLevel})
        \State $G(cr,cf) = \omega_1(cr)g_1(cr) + \omega_2(cl)g_2(cl)$
        \State $s_i \leftarrow $ apply (\ref{eq:speedUpdate})
        \EndFor
    \end{algorithmic}
\end{algorithm}
 
\section{Experimental Evaluation}
\label{sec:experimental}
In this section, our active sensing planning approach is evaluated via a simu-realistic pipeline by incorporating a high-fidelity simulator and a large-scale dataset for precision agriculture applications.  

\subsection{Dataset}
The exploited dataset was WeedMap~\cite{2018weedmap}, which contains multi-spectral images from sugar beet crops and weeds interfering in the crop lines. Data were collected during two campaigns, the first led to $3$ orthomosaic maps while the second to $5$. For every map the depicted plants were pixel-wise annotated, leading to 3 different classes, namely crop, weed and background. Every orthomosaic is provided also in a tiled version, where the original image is divided into patches of $480\times360$ pixels. In our case, only RGB data from the second campaign were utilized. 

\subsection{Detection Model}
Regarding the detection model $\cal{M}$ that semantically segments crop and weed instances, a deep-learning method was utilized. In specific, the well-known UNet~\cite{ronneberger2015u} architecture enhanced with EfficienNetB1~\cite{tan2019efficientnet} network as backbone was employed. This design was selected based on the balanced trade-off amongst inference time and model accuracy, taking into consideration that our aim was to deploy a real-time operating system. The deployed model was trained on WeedMap for $500$ epochs. A set of image processing techniques was utilized for data augmentation, in specific, image rotation, resize, vertical/horizontal flip and brightness change. At last, $255\times25$5 patches were randomly cropped from the tiled input images. Training was conducted with Adam optimizer with learning rate and batch size equal to $10^{-3}$ and $16$, respectively.

\subsection{Setup}
\label{subsec:setup}

To evaluate the proposed method in the context of the aforementioned dataset and assess its performance as a real-time interaction system, a hybrid simu-realistic framework was designed. The main goal here is to simulate real-world missions, with real-time interactions, in order to generate the required set of viewpoints $w_i$, collect the corresponding images $I_i$ and produce the most optimal field representation i.e a 2D orthomosaic, within the minimum operational time.

To accurately simulate the UAV's physics and dynamics and emulate its motion control, AirSim \cite{shah2018airsim}, an open-source high-fidelity simulator for autonomous vehicles, was utilized. AirSim is capable of forwarding the world dynamics, including a wide range of weather dynamics, at a high frequency allowing for real-time, hardware-in-the-loop ready, realistic simulations. All the experiments were carried out with a single drone within AirSim platform. The geo-referenced orthomosaic images of WeedMap fields, allow the direct mapping of the simulated UAV location in world coordinates to the pixel-level coordinates of the corresponding field's orthophoto. Thus, the exploited testing fields can be considered as natural parts of the environment and the UAV's camera input can be simulated by cropping image patches from the related orthophoto, with respect to the vehicle position. In Table \ref{table:path planning and sensor specifications} are provided further details regarding the parameters related to the UAV flight and the simulated camera sensor. The selection was based on the corresponding information provided in WeedMap dataset.

\begin{table}[ht!]
\caption{Path planning and sensor specifications.}
\label{table:path planning and sensor specifications}
\sisetup{table-format=2.4}
\centering
\small
\scalebox{0.8}{
\begin{tabular} {c | c c c } 
    \toprule
\textbf{Type} & \textbf{Description} & {\textbf{Specification} }  & {\textbf{Unit}} \\
\midrule
\multirow{2}{*}{UAV System} & Flight altitude & 10 & meters \\
& $f_{best}$ & 1 & frame/sec \\
\midrule
\multirow{3}{*}{Visual Sensor} 
& Overlap      & 70 &  \%  \\
& Gimbal pitch & -90 & degrees \\
& Image size (width $\times$ height) & $640 \times 480$ & pixels \\ 
\bottomrule
\end{tabular}}
\end{table}

\begin{figure*}[ht!]
    \centering
    \includegraphics[width=0.99\textwidth]{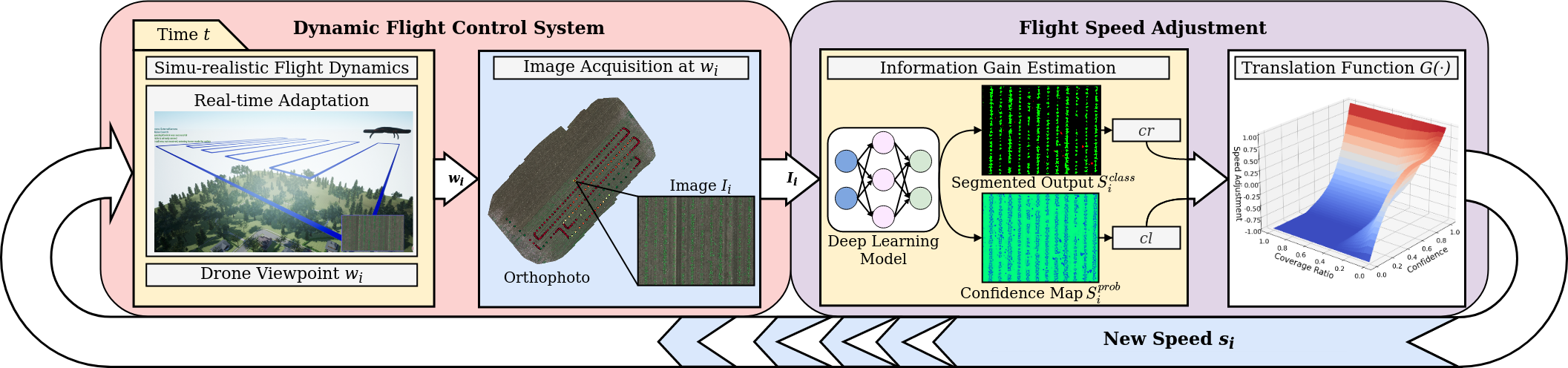}
    \caption{Graphical illustration demonstrating the core loop of the designed experimental setup. A simu-realistic flight environment, based on AirSim, is deployed to produce, in real time, UAV viewpoint $w_i$. Image $I_i$ is cropped at $w_i$ position from the field orthophoto and blurred according to current UAV speed. $I_i$ is processed by the proposed adaptive scheme to estimate the information gain of the scene and adjust UAV speed to $s_i$, based on the designed $G(\cdot)$ translation function.}
    \label{fig:pipeline}
\end{figure*}

 In this light, for each agricultural field in the deployed dataset, QGIS platform\footnote{\url{https://qgis.org/en/site/}} was used to specify the filled-in polygon ${\cal P}$ of (\ref{eq:stcKernel}) and an STC-based coverage path was designed and integrated into AirSim. During the simulated flight with initial speed $\bar{s}$, a set of processing operations are applied in a recursive manner in order to adapt the UAV speed in real-time. The core loop of this process is illustrated in figure~\ref{fig:pipeline}. In specific, with time interval $f_{best}$, the coordinates of drone viewpoint $w_i$ are extracted from AirSim environment. The acquired point is mapped to the corresponding geo-referenced orthomosaic image of the examined field and a $640 \times 480$ image is cropped according to the AirSim-emulated UAV trajectory. Furthermore, motion blur is applied to the cropped image according to the current UAV speed, aiming to create realistic captured data. In specific, we followed the formulation presented in \cite{fergus2006removing}. Assuming there is no additive noise, the blurred image $B_i$ is simply acquired by the convolution of a blur kernel $K$ with the captured image $I_i$, i.e $B_i =  K \ast I_i$. We know that the UAV is moving in the same direction as the vertical axis of the captured images. Thus, the blur kernel $K$ can be easily emulated with a vertical kernel (ones in the middle column and zeros everywhere else). In order to simulate the blurring effect impact according to vehicle speed, we increased the kernel size, e.g. $3 \times 3, 5 \times 5,$ etc, respectively. Next, the acquired image $I_i$ is forwarded to the proposed adaptive scheme, that assess the information gain enclosed in it and adapts the vehicle speed based on the proposed translation function $G(\cdot)$. The update information is fed back to the simu-realistic environment, regulating the UAV speed on-the-fly. The aforementioned process is repeated at the next time interval, for viewpoint $w_{i+1}$.

\subsection{Baseline}
To evaluate the efficiency of the proposed speed adjustment methodology, we chose to compare it against the ``go-to'' STC-based coverage path-planning approach for precision agriculture applications~\cite{cabreira2019survey, karatzinis2020towards, pham2017aerial}, where the UAV is moving with constant speed. Note that the flight path in both scenarios is identical, while the sampling interval remains the same in all cases. However, variations in speed lead to collecting data from different viewpoints $w_i$. Moreover, according to the vehicle speed during the capturing time, acquired images differ in terms of image quality due to motion blurring. We refer to the deployed non-adaptive method as \textit{STC-PA}, while the proposed method is mentioned as {\it OverFOMO}. Through this comparison, we aim to answer the following question: instead of covering the field with constant speed $\bar{s}$, can the speed adjustments of the proposed method lead to more meaningful data in less or comparative time?

\subsection{Performance Analysis}
\label{subsec:performance_analysis}
The proposed method was extensively evaluated under different flight scenarios and agricultural environments. More specifically, for each one of the 5 crop areas, we deployed the adaptive planning process through the aforementioned simurealistic pipeline, for different selections of nominal speed, in $m/s$, namely $\bar{s}\in{\{3, 4, 5, 6\}}$. $q$ parameter of (\ref{eq:speedUpdate}) was set to $1$ implying that UAV can increase or decrease its nominal speed by $1$ m/s at maximum. The evaluation process is based on recreating the orthomosaic map from the set of images $I$ collected during the {\it OverFOMO} mission. Next, the stitched outcome is semantically segmented, utilizing the aforementioned trained model, and IoU is calculated for crop and weed class. Our aim is to quantify the quality of the reconstructed map in terms of the enclosed semantic content and thus, provide a metric of the scanning efficiency of the planned mission.

\begin{figure}[ht!]
    \centering
    \includegraphics[width=0.95\columnwidth]{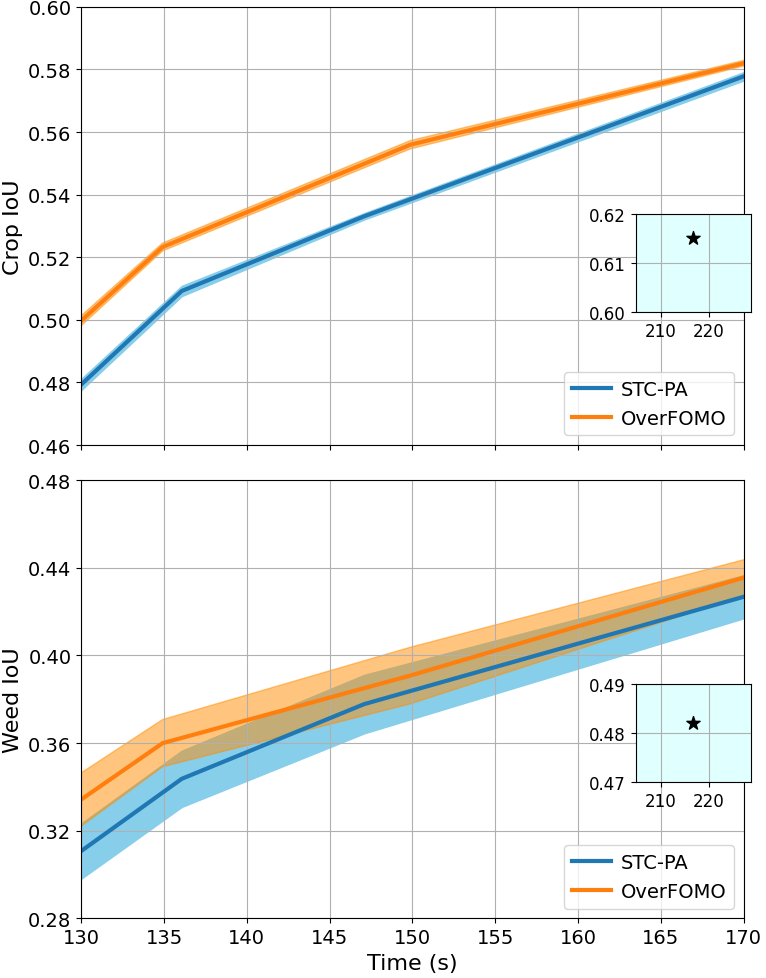}
    \caption{Averaged IoU for the testing fields of Weedmap dataset. Solid line refers to mean value and shaded region to variance. Black star refers to the ideal scenario where \textit{STC-PA} method is applied with nominal speed $\bar{s}=2$ $m/s$ and can be considered as the convergence point of the two methods. For both crop (top) and weed (bottom) classes, the proposed adaptive method leads to higher performance, implying more accurate scanning of the examined area.}
    \label{fig::iou_crop_weed}
\end{figure}

The aforementioned evaluation process is applied for each one of the examined fields, computing the execution time and the IoU for crop and weed class. In order to conduct credible validations, in each case the testing field is excluded from the training process of the detection model. The same approach is followed for both the \textit{STC-PA} and the {\it OverFOMO} approach. The two methods are compared in figure \ref{fig::iou_crop_weed}, where is presented the average IoU over the 5 examined fields and its variance for crop and weed class correspondingly, for different nominal speeds. In total, 20 flight scenarios (4 nominal speeds $\times$ 5 fields) were executed for each of the two evaluated approaches. Furthermore, for the \textit{STC-PA} method we examine the case of $\bar{s}=2$ $m/s$ which is considered as the ideal scenario, where the UAV is moving with the minimum speed and thus, data are collected totally undistorted (no motion blur is applied). 

For both classes, the proposed method outperforms \textit{STC-PA}. The efficiency of adaptive planning, in terms of IoU, is clear in case of crop detection, while in case of weed the maximum IoU of the proposed method is constantly higher than the comparative for the whole set of examined nominal speeds. In terms of execution time, for lower values of nominal speed, the adaptive method is to some extent slower yet, in favor of higher accuracy. As the nominal speed increases the execution time gap between the two methods is decreased, while for $\bar{s}=6$ $m/s$, the proposed method outperforms the \textit{STC-PA} in terms of flight time also. All in all, results imply that {\it OverFOMO} scans efficiently an examined area, collecting high quality data from the areas containing rich semantic content while passing by areas of lower interest to reduce the flight time.

\subsection{Qualitative Analysis}
\begin{figure*}[ht!]
    \centering
    \includegraphics[width=0.99\textwidth]{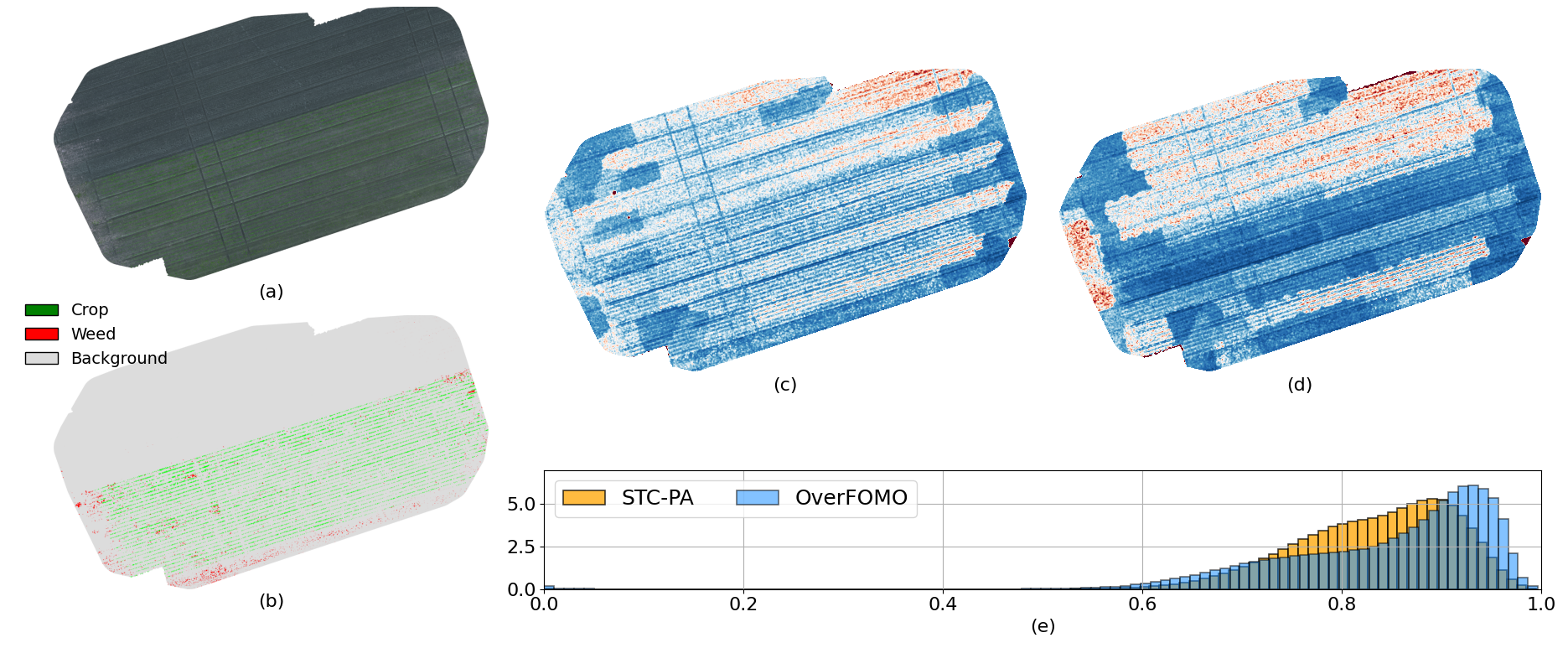}
    \caption{Qualitative results for the \textit{STC-PA} and the proposed {\it OverFOMO} approach. In (a) is presented the original orthomosaic image, while in (b) the semantic content of the examined scene. In (c) and (d) is illustrated the image similarity, in terms of SSIM, among the original (b) and the generated orthomosaic via coverage missions planned following the \textit{STC-PA} and the {\it OverFOMO} method, respectively. Both missions are conducted with same nominal speed. Red colors resemble lower values of SSIM, while higher values of SSIM are mapped with blue colors.In (e) is presented the corresponding histogram of the calculated SSIM index for both cases.}
    \label{fig:ssim_maps}
\end{figure*}

In order to validate further the efficiency of the developed method we evaluate the generated orthomosaic maps in terms of image quality. Towards this direction, simulated missions deployed with the \textit{STC-PA} and the {\it OverFOMO} method are conducted for the field ``002'' of the Weedmap dataset, with nominal speed $\bar{s}=3$ $m/s$. Next, we estimate the image similarity, in terms of Structural Similarity Index (SSIM)~\cite{wang2004image}, among the original orthomosaic (provided in the dataset) and the one built via data collected from the {\it OverFOMO} mission. The same process is followed for the \textit{STC-PA} method. In figure \ref{fig:ssim_maps} qualitative results for the two comparative approaches are presented. More specifically, in figure \ref{fig:ssim_maps}(a) the original orthomasaic image is presented, while in figure \ref{fig:ssim_maps}(b) is illustrated the annotated ground truth, aiming to provided further insights regarding the semantic content of the examined scene. In figure \ref{fig:ssim_maps}(c) \& (d) the estimated SSIM index is demonstrated for the cases of \textit{STC-PA} and {\it OverFOMO}, respectively. For visualization purposes, the similarity of the generated orthomosaics to the original one is illustrated in red-blue colorscale. Blue areas indicate higher similarity, while yellow and red regions indicate deviations between the generated and the original image. For a more comprehensive comparison, the histogram of the calculated SSIM values is provided for each case in figure \ref{fig:ssim_maps}(e).

Results imply that the proposed method leads to a more accurate orthomosaic map compared to the current ``go-to'' approach, especially for areas of high information gain, where the measured similarity is higher (note dark blue regions in figure \ref{fig:ssim_maps}(d)). The fidelity of the generated orthomosaic indicates that the collected data of the adaptive mission can enclose more precisely the semantic content of the scene. This is also supported by the provided histograms in figure \ref{fig:ssim_maps}(e), where the proposed method reports higher similarity values for the majority of cases. Taking into consideration the overall patterns of SSIM index values, with respect to the information of figure \ref{fig:ssim_maps}(b), one can derive that the proposed method regulates the vehicle speed according to the semantic content of the scanned field. In areas where the information gain is high, i.e. lush vegetation, UAV decelerates to acquire high quality - less blurry - data, while in areas of lower interest it accelerates since the information gain is considered minimum. On the contrary, the \textit{STC-PA} method of constant speed scanning presents, to some extent, constant image quality levels, distributed across the whole field, without taking into consideration the semantic content of the scene.

\section{Conclusions}
In this work an UAV active sensing coverage path planning scheme for precision agriculture tasks has been presented. Our method is capable of adjusting the UAV speed based on the perceived visual information (i.e. observed crops and weeds), while the computational needs for the online processing are uncoupled to the operation field's size. A core-element of the proposed approach is a robust deep learning-based module, allowing to regulate the vehicle speed according to the quantity of the detected instances and the quality (confidence) of such detections. The proposed method has been extensively validated through a designed simu-realistic environment, conducting several missions with different nominal speed for $5$ different agricultural fields of WeedMap dataset. Compared to the well-known lawn-mover coverage path planning, our method manages to capture higher quality data in comparable execution times. In the future we aim to deploy our method in real-world scenarios by employing UAVs with on-board capabilities.    
\label{sec:conclusions}





\section{Acknowledgment}
This research has been financed by the European Regional Development Fund of the European Union and Greek national funds through the Operational Program Competitiveness, Entrepreneurship and Innovation, under the call RESEARCH - CREATE - INNOVATE (T1EDK-00636). We gratefully acknowledge the support of NVIDIA Corporation with the donation of GPUs used for this research.

\appendix
\section{Image Quality vs Method Performance}

\begin{figure*}[ht!]
    \centering
    \includegraphics[width=0.99\textwidth]{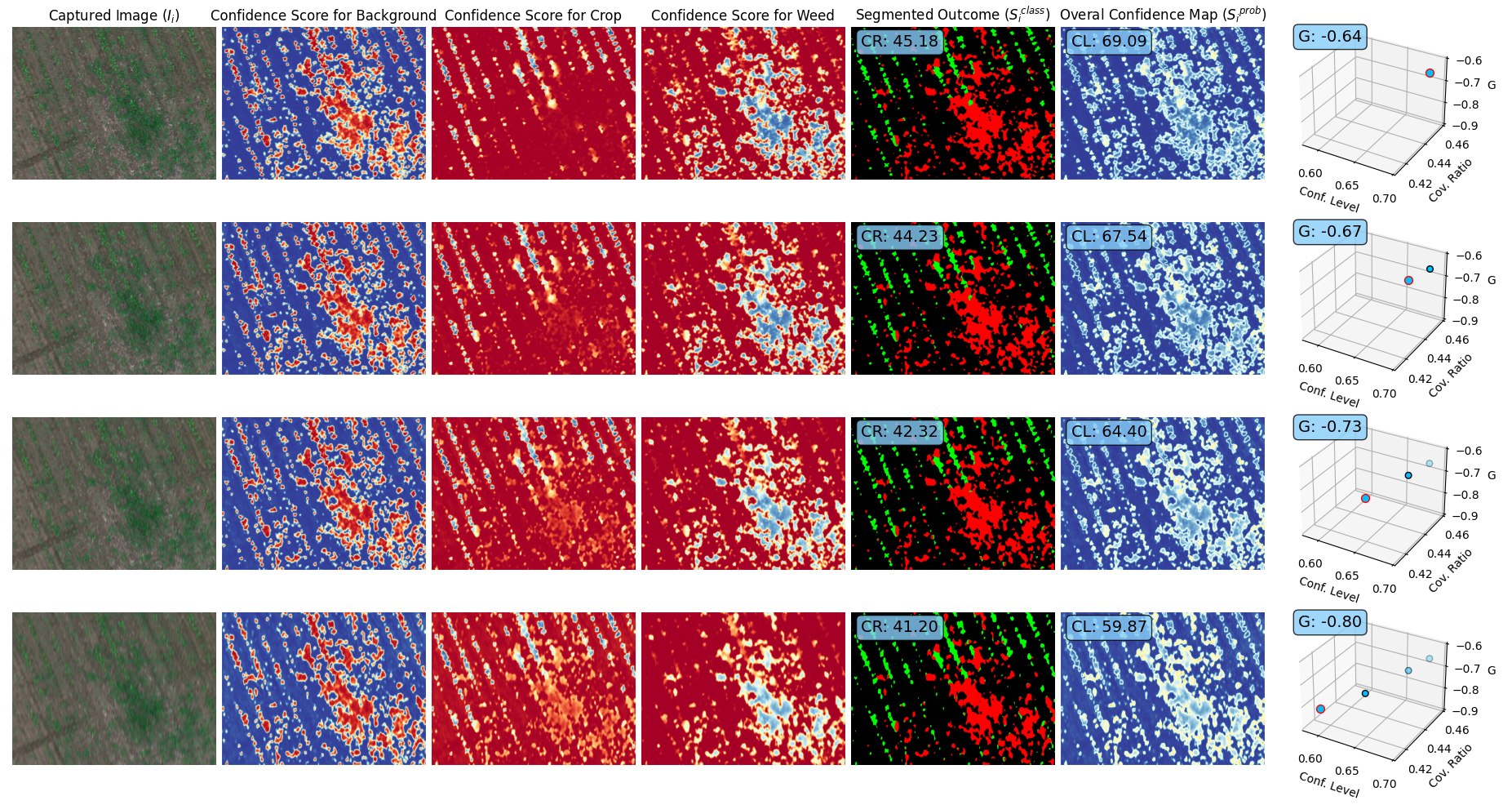}
    \caption{Illustration of the impact of motion blur on the proposed method performance. In each row, the motion blurring applied to the input image is increased, leading to lower values of $cl$ (\%) metric, although $cr$ (\%) remains at similar levels. Last column presents in the $3D$ space the calculated $G$ value (red outline) among with the corresponding values of the previous blurring cases. One can note the gradual decrease of $G$ value, implying the reduction of vehicle speed. All in all, the proposed OverFOMO approach takes into consideration the confidence of the segmentation model and adjusts the UAV speed accordingly to acquire more accurate estimations that meet the application-oriented requirements.}
    \label{fig:increase_blur}
\end{figure*}

In this appendix are presented further details regarding how the proposed method's performance is affected from the efficiency of the employed $\cal{M}$ model, the deduction of image quality due to speed increment and the possible misclassifications.

More specifically, the clarity of the on-the-fly captured images affects the segmentation confidence during the online phase of the adaptive system. Through the extensive evaluation of the deployed deep-learning model, we noticed that motion blur mostly affects the clearness of the depicted weeds and crops, increasing the ambiguity of their exact shape and size, and under the perspective of Bayesian modeling~\cite{kendall2017uncertainties}, increasing the \textit{aleatoric} uncertainty. \textit{Epistemic} uncertainty is also inherent in the prediction system, and it is reflected in the deviation of the captured image from the distribution of the training data. The offline validation of the employed semantic segmentation model, implied that it can generalize well in previously unseen data and thus, the effect of the epistemic uncertainty is not crucial. However, training data refer to an ideal scenario where the utilized images contain no distortion. Thus, during the online phase, aleatoric uncertainty expressed through the blurring effect significantly affects the efficiency of the on-the-fly prediction.

The above analysis comprises the challenging nature of the problem that we aim to tackle. Towards this direction, we use this uncertainty to our advantage in order to regulate the UAV speed according to it. By considering the confidence score of the segmented outcome, through the $cl$ metric, we aim to estimate the information gain at each sensing waypoint in respect to the confidence of this estimation.

In figure \ref{fig:increase_blur} we present the proposed method performance for an input image which is gradually deteriorated via motion blurring. One can notice that although the $cr$ metric is slightly decreased, the $cl$ value is significantly dropped, implying uncertainty in the acquired estimations and deterioration of the image quality due to the enhancement of the blurring effect. Please note the calculated $G$ values in all cases, which are gradually decreased, implying the adjustment of speed to lower values. The presented illustration demonstrates the ability of the proposed method to adapt the vehicle speed in order to avoid missing vital information and cope with possible misclassifications due to low image quality.

\section{Calibration of Translation and Weighting Functions}
\label{AppendixB}
In order to obtain the $g(\cdot)$ and $w(\cdot)$ functions of figure \ref{fig:functionsWeights} we followed a reverse engineering approach to make the adaptive system meet the expected behavior of the characteristic cases presented in figure \ref{fig:different_cases}. According to the presented formulation for $G(\cdot)$, we want the translation functions $g_1(\cdot), g_2(\cdot)$ to map the input to values from -1 to 1. Similarly, the weight functions $w_1(\cdot), w_2(\cdot)$ should range from 0 to 1 and sum to 1. Based on that, we focused on a family of linear-wise and parabola functions for $g(\cdot)$ and $w(\cdot)$, respectively. Moreover, during the training and evaluation of the $\cal{M}$ model that segments the images, we acquired valuable insights. First, we know that since it is a $3$ class problem, the probability $p_j$ cannot be lower than $0.33$. Thus, we do not expect values lower than that for $cl$ metric. Moreover, by examining sample images of the evaluation set we concluded that adequately accurate detections are acquired when $cl$ is around $0.75$, thus we considered this as a break-point. Similarly, we noticed that at the current altitude the peak coverage ratio is around $0.4$ while $cr$ values below $0.15$ refer to areas of low vegetation. Regarding the $w(\cdot)$ parabola functions, they were designed to control the contribution of each metric and express the system's expected behavior. We want to ignore the estimated coverage ratio in case that this estimation is ambiguous or overly strong. In case of moderate belief, we want the system to be guided accordingly, taking also into consideration the $cr$ value. Please note that the presented functions are not the unique solution, even for the specific IPP problem. One can select different functions, or tune their key-points according to the use-case and in respect to how much tolerance can be enclosed to the information quality - speed trade off.

\bibliographystyle{elsarticle-num-names} 
\bibliography{References}

\begin{thebibliography}{38}
\expandafter\ifx\csname natexlab\endcsname\relax\def\natexlab#1{#1}\fi
\providecommand{\url}[1]{\texttt{#1}}
\providecommand{\href}[2]{#2}
\providecommand{\path}[1]{#1}
\providecommand{\DOIprefix}{doi:}
\providecommand{\ArXivprefix}{arXiv:}
\providecommand{\URLprefix}{URL: }
\providecommand{\Pubmedprefix}{pmid:}
\providecommand{\doi}[1]{\href{http://dx.doi.org/#1}{\path{#1}}}
\providecommand{\Pubmed}[1]{\href{pmid:#1}{\path{#1}}}
\providecommand{\bibinfo}[2]{#2}
\ifx\xfnm\relax \def\xfnm[#1]{\unskip,\space#1}\fi
\bibitem[{Martinez-Alpiste et~al.(2021)Martinez-Alpiste, Golcarenarenji, Wang,
  and Alcaraz-Calero}]{martinez2021search}
\bibinfo{author}{I.~Martinez-Alpiste}, \bibinfo{author}{G.~Golcarenarenji},
  \bibinfo{author}{Q.~Wang}, \bibinfo{author}{J.~M. Alcaraz-Calero},
\newblock \bibinfo{title}{Search and rescue operation using uavs: A case
  study},
\newblock \bibinfo{journal}{Expert Systems with Applications}
  \bibinfo{volume}{178} (\bibinfo{year}{2021}) \bibinfo{pages}{114937}.
\bibitem[{Kim et~al.(2019)Kim, Liu, Lee, and Kamat}]{kim2019remote}
\bibinfo{author}{D.~Kim}, \bibinfo{author}{M.~Liu}, \bibinfo{author}{S.~Lee},
  \bibinfo{author}{V.~R. Kamat},
\newblock \bibinfo{title}{Remote proximity monitoring between mobile
  construction resources using camera-mounted uavs},
\newblock \bibinfo{journal}{Automation in Construction} \bibinfo{volume}{99}
  (\bibinfo{year}{2019}) \bibinfo{pages}{168--182}.
\bibitem[{Pham et~al.(2018)Pham, La, Feil-Seifer, and
  Deans}]{pham2018distributed}
\bibinfo{author}{H.~X. Pham}, \bibinfo{author}{H.~M. La},
  \bibinfo{author}{D.~Feil-Seifer}, \bibinfo{author}{M.~C. Deans},
\newblock \bibinfo{title}{A distributed control framework of multiple unmanned
  aerial vehicles for dynamic wildfire tracking},
\newblock \bibinfo{journal}{IEEE Transactions on Systems, Man, and Cybernetics:
  Systems} \bibinfo{volume}{50} (\bibinfo{year}{2018})
  \bibinfo{pages}{1537--1548}.
\bibitem[{Rodr{\'\i}guez et~al.(2021)Rodr{\'\i}guez, Lizarazo, Prieto, and
  Angulo-Morales}]{rodriguez2021assessment}
\bibinfo{author}{J.~Rodr{\'\i}guez}, \bibinfo{author}{I.~Lizarazo},
  \bibinfo{author}{F.~Prieto}, \bibinfo{author}{V.~Angulo-Morales},
\newblock \bibinfo{title}{Assessment of potato late blight from uav-based
  multispectral imagery},
\newblock \bibinfo{journal}{Computers and Electronics in Agriculture}
  \bibinfo{volume}{184} (\bibinfo{year}{2021}) \bibinfo{pages}{106061}.
\bibitem[{O’Mahony et~al.(2019)O’Mahony, Campbell, Carvalho, Harapanahalli,
  Hernandez, Krpalkova, Riordan, and Walsh}]{o2019deep}
\bibinfo{author}{N.~O’Mahony}, \bibinfo{author}{S.~Campbell},
  \bibinfo{author}{A.~Carvalho}, \bibinfo{author}{S.~Harapanahalli},
  \bibinfo{author}{G.~V. Hernandez}, \bibinfo{author}{L.~Krpalkova},
  \bibinfo{author}{D.~Riordan}, \bibinfo{author}{J.~Walsh},
\newblock \bibinfo{title}{Deep learning vs. traditional computer vision},
\newblock in: \bibinfo{booktitle}{Science and information conference},
  \bibinfo{organization}{Springer}, \bibinfo{year}{2019}, pp.
  \bibinfo{pages}{128--144}.
\bibitem[{Gonz{\'a}lez-Garc{\'\i}a et~al.(2020)Gonz{\'a}lez-Garc{\'\i}a,
  Swenson, and G{\'o}mez-Espinosa}]{gonzalez2020real}
\bibinfo{author}{J.~Gonz{\'a}lez-Garc{\'\i}a}, \bibinfo{author}{R.~L. Swenson},
  \bibinfo{author}{A.~G{\'o}mez-Espinosa},
\newblock \bibinfo{title}{Real-time kinematics applied at unmanned aerial
  vehicles positioning for orthophotography in precision agriculture},
\newblock \bibinfo{journal}{Computers and Electronics in Agriculture}
  \bibinfo{volume}{177} (\bibinfo{year}{2020}) \bibinfo{pages}{105695}.
\bibitem[{Ampatzidis et~al.(2020)Ampatzidis, Partel, and
  Costa}]{ampatzidis2020agroview}
\bibinfo{author}{Y.~Ampatzidis}, \bibinfo{author}{V.~Partel},
  \bibinfo{author}{L.~Costa},
\newblock \bibinfo{title}{Agroview: Cloud-based application to process, analyze
  and visualize uav-collected data for precision agriculture applications
  utilizing artificial intelligence},
\newblock \bibinfo{journal}{Computers and Electronics in Agriculture}
  \bibinfo{volume}{174} (\bibinfo{year}{2020}) \bibinfo{pages}{105457}.
\bibitem[{Karatzinis et~al.(2020)Karatzinis, Apostolidis, Kapoutsis,
  Panagiotopoulou, Boutalis, and Kosmatopoulos}]{karatzinis2020towards}
\bibinfo{author}{G.~D. Karatzinis}, \bibinfo{author}{S.~D. Apostolidis},
  \bibinfo{author}{A.~C. Kapoutsis}, \bibinfo{author}{L.~Panagiotopoulou},
  \bibinfo{author}{Y.~S. Boutalis}, \bibinfo{author}{E.~B. Kosmatopoulos},
\newblock \bibinfo{title}{Towards an integrated low-cost agricultural
  monitoring system with unmanned aircraft system},
\newblock in: \bibinfo{booktitle}{2020 International conference on unmanned
  aircraft systems (ICUAS)}, \bibinfo{organization}{IEEE},
  \bibinfo{year}{2020}, pp. \bibinfo{pages}{1131--1138}.
\bibitem[{Bah et~al.(2018)Bah, Dericquebourg, Hafiane, and
  Canals}]{bah2018deep}
\bibinfo{author}{M.~D. Bah}, \bibinfo{author}{E.~Dericquebourg},
  \bibinfo{author}{A.~Hafiane}, \bibinfo{author}{R.~Canals},
\newblock \bibinfo{title}{Deep learning based classification system for
  identifying weeds using high-resolution uav imagery},
\newblock in: \bibinfo{booktitle}{Science and Information Conference},
  \bibinfo{organization}{Springer}, \bibinfo{year}{2018}, pp.
  \bibinfo{pages}{176--187}.
\bibitem[{R{\"u}ckin et~al.(2022)R{\"u}ckin, Jin, Magistri, Stachniss, and
  Popovi{\'c}}]{ruckin2022informative}
\bibinfo{author}{J.~R{\"u}ckin}, \bibinfo{author}{L.~Jin},
  \bibinfo{author}{F.~Magistri}, \bibinfo{author}{C.~Stachniss},
  \bibinfo{author}{M.~Popovi{\'c}},
\newblock \bibinfo{title}{Informative path planning for active learning in
  aerial semantic mapping},
\newblock \bibinfo{journal}{arXiv preprint arXiv:2203.01652}
  (\bibinfo{year}{2022}).
\bibitem[{Meliou et~al.(2007)Meliou, Krause, Guestrin, and
  Hellerstein}]{meliou2007nonmyopic}
\bibinfo{author}{A.~Meliou}, \bibinfo{author}{A.~Krause},
  \bibinfo{author}{C.~Guestrin}, \bibinfo{author}{J.~M. Hellerstein},
\newblock \bibinfo{title}{Nonmyopic informative path planning in
  spatio-temporal models},
\newblock in: \bibinfo{booktitle}{AAAI}, volume~\bibinfo{volume}{10},
  \bibinfo{year}{2007}, pp. \bibinfo{pages}{16--7}.
\bibitem[{Gabriely and Rimon(2001)}]{gabriely2001spanning}
\bibinfo{author}{Y.~Gabriely}, \bibinfo{author}{E.~Rimon},
\newblock \bibinfo{title}{Spanning-tree based coverage of continuous areas by a
  mobile robot},
\newblock \bibinfo{journal}{Annals of mathematics and artificial intelligence}
  \bibinfo{volume}{31} (\bibinfo{year}{2001}) \bibinfo{pages}{77--98}.
\bibitem[{Choset and Pignon(1998)}]{choset1998coverage}
\bibinfo{author}{H.~Choset}, \bibinfo{author}{P.~Pignon},
\newblock \bibinfo{title}{Coverage path planning: The boustrophedon cellular
  decomposition},
\newblock in: \bibinfo{booktitle}{Field and service robotics},
  \bibinfo{organization}{Springer}, \bibinfo{year}{1998}, pp.
  \bibinfo{pages}{203--209}.
\bibitem[{Cabreira et~al.(2019)Cabreira, Brisolara, and
  Paulo~R}]{cabreira2019survey}
\bibinfo{author}{T.~M. Cabreira}, \bibinfo{author}{L.~B. Brisolara},
  \bibinfo{author}{F.~J. Paulo~R},
\newblock \bibinfo{title}{Survey on coverage path planning with unmanned aerial
  vehicles},
\newblock \bibinfo{journal}{Drones} \bibinfo{volume}{3} (\bibinfo{year}{2019})
  \bibinfo{pages}{4}.
\bibitem[{Williams and Rasmussen(2006)}]{williams2006gaussian}
\bibinfo{author}{C.~K. Williams}, \bibinfo{author}{C.~E. Rasmussen},
  \bibinfo{title}{Gaussian processes for machine learning},
  volume~\bibinfo{volume}{2}, \bibinfo{publisher}{MIT press Cambridge, MA},
  \bibinfo{year}{2006}.
\bibitem[{Hitz et~al.(2017)Hitz, Galceran, Garneau, Pomerleau, and
  Siegwart}]{hitz2017adaptive}
\bibinfo{author}{G.~Hitz}, \bibinfo{author}{E.~Galceran},
  \bibinfo{author}{M.-{\`E}. Garneau}, \bibinfo{author}{F.~Pomerleau},
  \bibinfo{author}{R.~Siegwart},
\newblock \bibinfo{title}{Adaptive continuous-space informative path planning
  for online environmental monitoring},
\newblock \bibinfo{journal}{Journal of Field Robotics} \bibinfo{volume}{34}
  (\bibinfo{year}{2017}) \bibinfo{pages}{1427--1449}.
\bibitem[{Vivaldini et~al.(2019)Vivaldini, Martinelli, Guizilini, Souza,
  Oliveira, Ramos, and Wolf}]{vivaldini2019uav}
\bibinfo{author}{K.~C. Vivaldini}, \bibinfo{author}{T.~H. Martinelli},
  \bibinfo{author}{V.~C. Guizilini}, \bibinfo{author}{J.~R. Souza},
  \bibinfo{author}{M.~D. Oliveira}, \bibinfo{author}{F.~T. Ramos},
  \bibinfo{author}{D.~F. Wolf},
\newblock \bibinfo{title}{Uav route planning for active disease
  classification},
\newblock \bibinfo{journal}{Autonomous robots} \bibinfo{volume}{43}
  (\bibinfo{year}{2019}) \bibinfo{pages}{1137--1153}.
\bibitem[{Popovi{\'c} et~al.(2017)Popovi{\'c}, Vidal-Calleja, Hitz, Sa,
  Siegwart, and Nieto}]{popovic2017multiresolution}
\bibinfo{author}{M.~Popovi{\'c}}, \bibinfo{author}{T.~Vidal-Calleja},
  \bibinfo{author}{G.~Hitz}, \bibinfo{author}{I.~Sa},
  \bibinfo{author}{R.~Siegwart}, \bibinfo{author}{J.~Nieto},
\newblock \bibinfo{title}{Multiresolution mapping and informative path planning
  for uav-based terrain monitoring},
\newblock in: \bibinfo{booktitle}{2017 IEEE/RSJ International Conference on
  Intelligent Robots and Systems (IROS)}, \bibinfo{organization}{IEEE},
  \bibinfo{year}{2017}, pp. \bibinfo{pages}{1382--1388}.
\bibitem[{Popovi{\'c} et~al.(2020)Popovi{\'c}, Vidal-Calleja, Chung, Nieto, and
  Siegwart}]{popovic2020informative}
\bibinfo{author}{M.~Popovi{\'c}}, \bibinfo{author}{T.~Vidal-Calleja},
  \bibinfo{author}{J.~J. Chung}, \bibinfo{author}{J.~Nieto},
  \bibinfo{author}{R.~Siegwart},
\newblock \bibinfo{title}{Informative path planning for active field mapping
  under localization uncertainty},
\newblock in: \bibinfo{booktitle}{2020 IEEE International Conference on
  Robotics and Automation (ICRA)}, \bibinfo{organization}{IEEE},
  \bibinfo{year}{2020}, pp. \bibinfo{pages}{10751--10757}.
\bibitem[{Stache et~al.(2023)Stache, Westheider, Magistri, Stachniss, and
  Popović}]{stache2021adaptive}
\bibinfo{author}{F.~Stache}, \bibinfo{author}{J.~Westheider},
  \bibinfo{author}{F.~Magistri}, \bibinfo{author}{C.~Stachniss},
  \bibinfo{author}{M.~Popović},
\newblock \bibinfo{title}{Adaptive path planning for uavs for multi-resolution
  semantic segmentation},
\newblock \bibinfo{journal}{Robotics and Autonomous Systems}
  \bibinfo{volume}{159} (\bibinfo{year}{2023}) \bibinfo{pages}{104288}.
  \URLprefix
  \url{https://www.sciencedirect.com/science/article/pii/S0921889022001774}.
  \DOIprefix\doi{https://doi.org/10.1016/j.robot.2022.104288}.
\bibitem[{Koutras et~al.(2020)Koutras, Kapoutsis, and
  Kosmatopoulos}]{koutras2020autonomous}
\bibinfo{author}{D.~I. Koutras}, \bibinfo{author}{A.~C. Kapoutsis},
  \bibinfo{author}{E.~B. Kosmatopoulos},
\newblock \bibinfo{title}{Autonomous and cooperative design of the monitor
  positions for a team of uavs to maximize the quantity and quality of detected
  objects},
\newblock \bibinfo{journal}{IEEE Robotics and Automation Letters}
  \bibinfo{volume}{5} (\bibinfo{year}{2020}) \bibinfo{pages}{4986--4993}.
\bibitem[{R{\"u}ckin et~al.(2022)R{\"u}ckin, Jin, and
  Popovi{\'c}}]{ruckin2021adaptive}
\bibinfo{author}{J.~R{\"u}ckin}, \bibinfo{author}{L.~Jin},
  \bibinfo{author}{M.~Popovi{\'c}},
\newblock \bibinfo{title}{Adaptive informative path planning using deep
  reinforcement learning for uav-based active sensing},
\newblock in: \bibinfo{booktitle}{2022 International Conference on Robotics and
  Automation (ICRA)}, \bibinfo{organization}{IEEE}, \bibinfo{year}{2022}, pp.
  \bibinfo{pages}{4473--4479}.
\bibitem[{Kapoutsis et~al.(2016)Kapoutsis, Chatzichristofis, Doitsidis,
  de~Sousa, Pinto, Braga, and Kosmatopoulos}]{kapoutsis2016real}
\bibinfo{author}{A.~C. Kapoutsis}, \bibinfo{author}{S.~A. Chatzichristofis},
  \bibinfo{author}{L.~Doitsidis}, \bibinfo{author}{J.~B. de~Sousa},
  \bibinfo{author}{J.~Pinto}, \bibinfo{author}{J.~Braga},
  \bibinfo{author}{E.~B. Kosmatopoulos},
\newblock \bibinfo{title}{Real-time adaptive multi-robot exploration with
  application to underwater map construction},
\newblock \bibinfo{journal}{Autonomous robots} \bibinfo{volume}{40}
  (\bibinfo{year}{2016}) \bibinfo{pages}{987--1015}.
\bibitem[{Renzaglia et~al.(2012)Renzaglia, Doitsidis, Martinelli, and
  Kosmatopoulos}]{renzaglia2012multi}
\bibinfo{author}{A.~Renzaglia}, \bibinfo{author}{L.~Doitsidis},
  \bibinfo{author}{A.~Martinelli}, \bibinfo{author}{E.~B. Kosmatopoulos},
\newblock \bibinfo{title}{Multi-robot three-dimensional coverage of unknown
  areas},
\newblock \bibinfo{journal}{The International Journal of Robotics Research}
  \bibinfo{volume}{31} (\bibinfo{year}{2012}) \bibinfo{pages}{738--752}.
\bibitem[{Shah et~al.(2018)Shah, Dey, Lovett, and Kapoor}]{shah2018airsim}
\bibinfo{author}{S.~Shah}, \bibinfo{author}{D.~Dey},
  \bibinfo{author}{C.~Lovett}, \bibinfo{author}{A.~Kapoor},
\newblock \bibinfo{title}{Airsim: High-fidelity visual and physical simulation
  for autonomous vehicles},
\newblock in: \bibinfo{booktitle}{Field and service robotics},
  \bibinfo{organization}{Springer}, \bibinfo{year}{2018}, pp.
  \bibinfo{pages}{621--635}.
\bibitem[{Pham et~al.(2017)Pham, Bestaoui, and Mammar}]{pham2017aerial}
\bibinfo{author}{T.~H. Pham}, \bibinfo{author}{Y.~Bestaoui},
  \bibinfo{author}{S.~Mammar},
\newblock \bibinfo{title}{Aerial robot coverage path planning approach with
  concave obstacles in precision agriculture},
\newblock in: \bibinfo{booktitle}{2017 Workshop on Research, Education and
  Development of Unmanned Aerial Systems (RED-UAS)},
  \bibinfo{organization}{IEEE}, \bibinfo{year}{2017}, pp.
  \bibinfo{pages}{43--48}.
\bibitem[{Sa et~al.(2018)Sa, Popovi{\'c}, Khanna, Chen, Lottes, Liebisch,
  Nieto, Stachniss, Walter, and Siegwart}]{2018weedmap}
\bibinfo{author}{I.~Sa}, \bibinfo{author}{M.~Popovi{\'c}},
  \bibinfo{author}{R.~Khanna}, \bibinfo{author}{Z.~Chen},
  \bibinfo{author}{P.~Lottes}, \bibinfo{author}{F.~Liebisch},
  \bibinfo{author}{J.~Nieto}, \bibinfo{author}{C.~Stachniss},
  \bibinfo{author}{A.~Walter}, \bibinfo{author}{R.~Siegwart},
\newblock \bibinfo{title}{Weedmap: A large-scale semantic weed mapping
  framework using aerial multispectral imaging and deep neural network for
  precision farming},
\newblock \bibinfo{journal}{Remote Sensing} \bibinfo{volume}{10}
  (\bibinfo{year}{2018}) \bibinfo{pages}{1423}.
\bibitem[{Krestenitis et~al.(2022)Krestenitis, Raptis, Kapoutsis, Ioannidis,
  Kosmatopoulos, Vrochidis, and Kompatsiaris}]{krestenitis2022cofly}
\bibinfo{author}{M.~Krestenitis}, \bibinfo{author}{E.~K. Raptis},
  \bibinfo{author}{A.~C. Kapoutsis}, \bibinfo{author}{K.~Ioannidis},
  \bibinfo{author}{E.~B. Kosmatopoulos}, \bibinfo{author}{S.~Vrochidis},
  \bibinfo{author}{I.~Kompatsiaris},
\newblock \bibinfo{title}{Cofly-weeddb: A uav image dataset for weed detection
  and species identification},
\newblock \bibinfo{journal}{Data in Brief} \bibinfo{volume}{45}
  (\bibinfo{year}{2022}) \bibinfo{pages}{108575}.
\bibitem[{Radoglou-Grammatikis et~al.(2020)Radoglou-Grammatikis, Sarigiannidis,
  Lagkas, and Moscholios}]{radoglou2020compilation}
\bibinfo{author}{P.~Radoglou-Grammatikis}, \bibinfo{author}{P.~Sarigiannidis},
  \bibinfo{author}{T.~Lagkas}, \bibinfo{author}{I.~Moscholios},
\newblock \bibinfo{title}{A compilation of uav applications for precision
  agriculture},
\newblock \bibinfo{journal}{Computer Networks} \bibinfo{volume}{172}
  (\bibinfo{year}{2020}) \bibinfo{pages}{107148}.
\bibitem[{Rezatofighi et~al.(2019)Rezatofighi, Tsoi, Gwak, Sadeghian, Reid, and
  Savarese}]{rezatofighi2019generalized}
\bibinfo{author}{H.~Rezatofighi}, \bibinfo{author}{N.~Tsoi},
  \bibinfo{author}{J.~Gwak}, \bibinfo{author}{A.~Sadeghian},
  \bibinfo{author}{I.~Reid}, \bibinfo{author}{S.~Savarese},
\newblock \bibinfo{title}{Generalized intersection over union: A metric and a
  loss for bounding box regression},
\newblock in: \bibinfo{booktitle}{Proceedings of the IEEE/CVF conference on
  computer vision and pattern recognition}, \bibinfo{year}{2019}, pp.
  \bibinfo{pages}{658--666}.
\bibitem[{Galceran and Carreras(2013)}]{galceran2013survey}
\bibinfo{author}{E.~Galceran}, \bibinfo{author}{M.~Carreras},
\newblock \bibinfo{title}{A survey on coverage path planning for robotics},
\newblock \bibinfo{journal}{Robotics and Autonomous systems}
  \bibinfo{volume}{61} (\bibinfo{year}{2013}) \bibinfo{pages}{1258--1276}.
\bibitem[{Apostolidis et~al.(2022)Apostolidis, Kapoutsis, Kapoutsis, and
  Kosmatopoulos}]{apostolidis2022cooperative}
\bibinfo{author}{S.~D. Apostolidis}, \bibinfo{author}{P.~C. Kapoutsis},
  \bibinfo{author}{A.~C. Kapoutsis}, \bibinfo{author}{E.~B. Kosmatopoulos},
\newblock \bibinfo{title}{Cooperative multi-uav coverage mission planning
  platform for remote sensing applications},
\newblock \bibinfo{journal}{Autonomous Robots}  (\bibinfo{year}{2022})
  \bibinfo{pages}{1--28}.
\bibitem[{Tsouros et~al.(2019)Tsouros, Bibi, and
  Sarigiannidis}]{tsouros2019review}
\bibinfo{author}{D.~C. Tsouros}, \bibinfo{author}{S.~Bibi},
  \bibinfo{author}{P.~G. Sarigiannidis},
\newblock \bibinfo{title}{A review on uav-based applications for precision
  agriculture},
\newblock \bibinfo{journal}{Information} \bibinfo{volume}{10}
  (\bibinfo{year}{2019}) \bibinfo{pages}{349}.
\bibitem[{Ronneberger et~al.(2015)Ronneberger, Fischer, and
  Brox}]{ronneberger2015u}
\bibinfo{author}{O.~Ronneberger}, \bibinfo{author}{P.~Fischer},
  \bibinfo{author}{T.~Brox},
\newblock \bibinfo{title}{U-net: Convolutional networks for biomedical image
  segmentation},
\newblock in: \bibinfo{booktitle}{International Conference on Medical image
  computing and computer-assisted intervention},
  \bibinfo{organization}{Springer}, \bibinfo{year}{2015}, pp.
  \bibinfo{pages}{234--241}.
\bibitem[{Tan and Le(2019)}]{tan2019efficientnet}
\bibinfo{author}{M.~Tan}, \bibinfo{author}{Q.~Le},
\newblock \bibinfo{title}{Efficientnet: Rethinking model scaling for
  convolutional neural networks},
\newblock in: \bibinfo{booktitle}{International conference on machine
  learning}, \bibinfo{organization}{PMLR}, \bibinfo{year}{2019}, pp.
  \bibinfo{pages}{6105--6114}.
\bibitem[{Fergus et~al.(2006)Fergus, Singh, Hertzmann, Roweis, and
  Freeman}]{fergus2006removing}
\bibinfo{author}{R.~Fergus}, \bibinfo{author}{B.~Singh},
  \bibinfo{author}{A.~Hertzmann}, \bibinfo{author}{S.~T. Roweis},
  \bibinfo{author}{W.~T. Freeman},
\newblock \bibinfo{title}{Removing camera shake from a single photograph},
\newblock in: \bibinfo{booktitle}{Acm Siggraph 2006 Papers},
  \bibinfo{year}{2006}, pp. \bibinfo{pages}{787--794}.
\bibitem[{Wang et~al.(2004)Wang, Bovik, Sheikh, and Simoncelli}]{wang2004image}
\bibinfo{author}{Z.~Wang}, \bibinfo{author}{A.~C. Bovik},
  \bibinfo{author}{H.~R. Sheikh}, \bibinfo{author}{E.~P. Simoncelli},
\newblock \bibinfo{title}{Image quality assessment: from error visibility to
  structural similarity},
\newblock \bibinfo{journal}{IEEE transactions on image processing}
  \bibinfo{volume}{13} (\bibinfo{year}{2004}) \bibinfo{pages}{600--612}.
\bibitem[{Kendall and Gal(2017)}]{kendall2017uncertainties}
\bibinfo{author}{A.~Kendall}, \bibinfo{author}{Y.~Gal},
\newblock \bibinfo{title}{What uncertainties do we need in bayesian deep
  learning for computer vision?},
\newblock \bibinfo{journal}{Advances in neural information processing systems}
  \bibinfo{volume}{30} (\bibinfo{year}{2017}).

\end{thebibliography}







\end{document}